\documentclass[10pt,twocolumn,letterpaper]{article}

\usepackage[pagenumbers]{cvpr} % To force page numbers, e.g. for an arXiv version

\usepackage[table]{xcolor} % xcolor is recommended for color definitions
\usepackage{colortbl}     % Provides \cellcolor
\usepackage{tikz}
\usepackage{pgfplots}
\usepackage{multirow}
\pgfplotsset{compat=1.18}

\newcommand{\yb}[1]{\cellcolor{yellow!35}\textbf{#1}}
\newcommand{\rb}[1]{\cellcolor{red!15}#1}

\newcommand{\tikzcheckcross}{%
    \tikz[scale=0.23] {
        \draw[line width=1,line cap=round,red!90!black] (0,0) -- (1,1);
        \draw[line width=1,line cap=round,red!90!black] (0,1) -- (1,0);
    }%
}
\newcommand{\checkcross}{This is the original content.}
\renewcommand{\checkcross}{\tikzcheckcross}
\newcommand{\tikzcheckmark}{%
    \tikz[scale=0.4] {
    \draw[line width=1,line cap=round,green!80!black] (0.25,0) -- (0.8,0.8);
    \draw[line width=1,line cap=round,green!80!black] (0,0.35) -- (0.23,0);
    }%
}
\newcommand{\checkmarkk}{This is the original content.}
\renewcommand{\checkmarkk}{\tikzcheckmark}

\newcommand{\smallfilledcircle}[1][0.5ex]{%
    \tikz \fill[green!80!black] (0.5,0.5) circle (#1);%
}
\newcommand{\circc}{This is the original content.}
\renewcommand{\circc}{\smallfilledcircle[0.5ex]}

\definecolor{cvprblue}{rgb}{0.21,0.49,0.74}
\usepackage[pagebackref,breaklinks,colorlinks,allcolors=cvprblue]{hyperref}

\def\paperID{*****} % *** Enter the Paper ID here
\def\confName{CVPR}
\def\confYear{2026}

\title{Edit-aware RAW Reconstruction}

\author{\\[-20pt] 
Abhijith Punnappurath \hspace{25pt} Luxi Zhao\textsuperscript{*} \hspace{25pt} Ke Zhao\textsuperscript{*} \\
Hue Nguyen \hspace{25pt} Radek Grzeszczuk \hspace{25pt} Michael S. Brown \\
AI Center–Toronto, Samsung Electronics \\
{\tt\small \{abhijith.p, lucy.zhao, k.zhao, h.nguyen1, radek.g, michael.b1\}@samsung.com}\\%[-20pt]  
}

\begin{document}
\maketitle

\def\thefootnote{*}\footnotetext{Equal contribution.}

\begin{abstract}
Users frequently edit camera images post-capture to achieve their preferred photofinishing style. While editing in the RAW domain provides greater accuracy and flexibility, most edits are performed on the camera’s display-referred output (e.g., 8-bit sRGB JPEG) since RAW images are rarely stored. Existing RAW reconstruction methods can recover RAW data from sRGB images, but these approaches are typically optimized for pixel-wise RAW reconstruction fidelity and tend to degrade under diverse rendering styles and editing operations.  We introduce a plug-and-play, edit-aware loss function that can be integrated into any existing RAW reconstruction framework to make the recovered RAWs more robust to different rendering styles and edits. Our loss formulation incorporates a modular, differentiable image signal processor (ISP) that simulates realistic photofinishing pipelines with tunable parameters. During training, parameters for each ISP module are randomly sampled from carefully designed distributions that model practical variations in real camera processing. The loss is then computed in sRGB space between ground-truth and reconstructed RAWs rendered through this differentiable ISP. Incorporating our loss improves sRGB reconstruction quality by up to 1.5–2 dB PSNR across various editing conditions. Moreover, when applied to metadata-assisted RAW reconstruction methods, our approach enables fine-tuning for target edits, yielding further gains. Since photographic editing is the primary motivation for RAW reconstruction in consumer imaging, our simple yet effective loss function provides a general mechanism for enhancing edit fidelity and rendering flexibility across existing methods.
\end{abstract}

%-------------------------------------------------------------------------

\begin{figure}
  \centering
  \includegraphics[width=\linewidth]{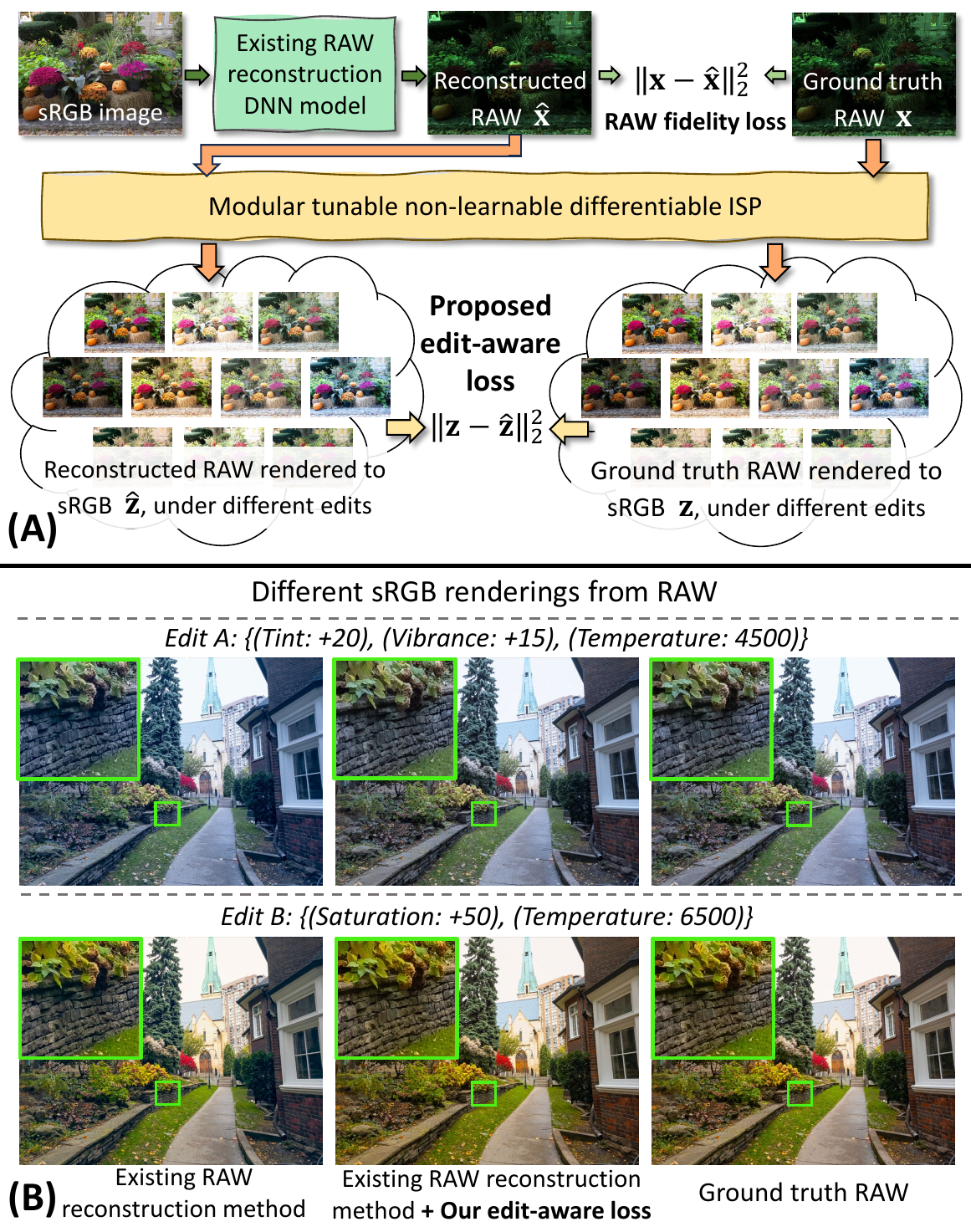}
  \caption{(A) Existing RAW reconstruction methods typically employ loss functions focused on accurate per-pixel RAW recovery (green path). However, the primary use of a reconstructed RAW image is to allow high-quality editing targeting sRGB outputs. This paper introduces an edit-aware loss for RAW reconstruction that yields higher sRGB fidelity across diverse edits (yellow path). (B) Two example edits applied to the RAW estimated by a conventional reconstruction method~\cite{rawdiff} (left), the RAW reconstructed by augmenting~\cite{rawdiff} with our edit-aware loss (middle), and the ground-truth RAW (right). Renderings are produced in Adobe Photoshop using the indicated settings. Our results more closely match the ground truth across edits.  
  }\label{fig:teaser}
%  \vspace{-5mm}
\end{figure}

\section{Introduction}
\label{sec:intro}

% \noindent Para one: Why do users edit images post-capture? 
Users edit camera-captured images in their photo galleries to achieve a preferred photofinishing style. Gallery images are the output of cameras' on-board image signal processor (ISP), which renders the RAW sensor image into a final photofinished output image~\cite{delbracio2021mobile}. The ISP follows a default rendering pipeline governed by fixed or discretely selectable photofinishing styles determined at capture time. Users often perform post-capture editing because they prefer a different style than the ISP default, or find that the ISP’s automatically selected settings produced a photofinished image that did not match their intent.

% \noindent Para two: What's wrong with editing the sRGB?
Post-capture editing in the RAW domain offers significant advantages over manipulating the photofinished image~\cite{kim2012new}. The RAW image represents the minimally processed sensor measurement of the physical scene, maintaining a linear response to scene radiance and preserving higher bit depth and dynamic range. In contrast, the ISP applies a series of operations---such as denoising, demosaicing, exposure and white balance correction, and color and tone mapping---to produce the final rendered output~\cite{delbracio2021mobile}. Many of these operations are non-invertible and introduce information loss through tonal compression, clipping, and quantization. Consequently, the photofinished image—--typically encoded in a perceptual color space (e.g., sRGB or Display P3) with 8–10 bits per channel and compressed using formats like JPEG or HEIC—--contains reduced information content. In consumer photography, the most common output remains 8-bit sRGB JPEG. Although editing RAW images allows more accurate and flexible adjustments due to their higher bit depth and tonal range, RAW files are rarely retained because of their large size and limited compatibility with standard display and sharing workflows.

% \noindent Para three: How do existing RAW reconstruction methods work?
RAW reconstruction methods aim to recover RAW sensor measurements from their rendered sRGB counterparts. Existing approaches can be broadly categorized into metadata-assisted and blind methods. Metadata-assisted approaches leverage additional capture-time information such as RAW samples~\cite{cam,sam,li2023metadata} or RAW latent vectors~\cite{r2lcm}, and typically achieve higher reconstruction fidelity than blind methods~\cite{rawdiff,cycleisp,invisp}, which rely solely on the sRGB input. The vast majority of RAW recovery methods optimize exclusively for pixel-level RAW reconstruction accuracy, without considering downstream use cases such as photographic rendering or editing. A few approaches~\cite{cycleisp,invisp} include cyclic losses that reprocess the recovered RAW to verify that it matches the unedited sRGB input image. While the fidelity loss is in the sRGB domain, such supervision does not provide robustness to diverse photofinishing styles or post-capture edits. Some methods (e.g.,~\cite{reraw}) have explored task-specific RAW reconstruction for computer vision applications (e.g., object detection), but these objectives are not aligned with the goals of photographic editing. In consumer photography, the primary motivation for RAW reconstruction is to enable high-quality, flexible image editing---a setting in which existing RAW-fidelity-driven approaches may produce suboptimal results. See the example in the first column of Fig.~\ref{fig:teaser}(B).

% To bridge this gap, we introduce an edit-aware framework that can be seamlessly integrated with existing RAW reconstruction methods, enabling them to better support diverse photographic rendering styles and editing workflows with higher fidelity and robustness.

\noindent \textbf{Contributions:}~
The limitations discussed above highlight the need for a reconstruction objective that directly accounts for the downstream goal of robust RAW-image editing supporting a wide variety of sRGB renderings. To address this gap, we introduce a plug-and-play edit-aware loss that can be integrated into any RAW reconstruction framework to improve robustness across a wide range of editing conditions. Our loss is formulated as a modular and tunable differentiable ISP that emulates realistic photofinishing pipelines by rendering RAW inputs to sRGB under a wide range of sampled edit settings. Importantly, our method does not assume access to the camera ISP that produced the input sRGB image. In practice, camera ISPs are proprietary black-box pipelines. Our approach instead uses a simplified differentiable ISP only as a training-time loss, without requiring knowledge of or access to the actual camera ISP. During training, parameters for each ISP module are randomly drawn from distributions designed to reflect common operating ranges, ensuring realistic variability. The loss is computed in sRGB space between the ground-truth and reconstructed RAW images, with both rendered through our differentiable ISP using the same random settings. See Fig.~\ref{fig:teaser}(A). By incorporating our edit-aware loss, the reconstructed RAW is optimized for editability and rendering robustness, trading a small amount of pixel-wise RAW-space fidelity for significantly improved sRGB rendering accuracy and consistency across edits.
% ---an intended and beneficial consequence of our design. 

We evaluate our approach across three representative RAW reconstruction models from both metadata-assisted and blind categories. We demonstrate that our loss generalizes to real-world editing workflows by applying diverse photographic edits in Adobe Photoshop. The introduction of our edit-aware loss, implemented through our lightweight differentiable ISP, consistently improves performance across all tested settings and we observe gains of up to 1.5--2 dB PSNR in sRGB space under various edits. An example is shown in the second column of Fig.~\ref{fig:teaser}(B). Overall, our approach provides a simple yet powerful mechanism for enhancing the practical utility of RAW reconstruction methods in real-world photographic editing scenarios.

%-------------------------------------------------------------------------

\section{Related work}
\label{sec:related}

\begin{figure*}[!t]
  \centering
  \includegraphics[width=\linewidth]{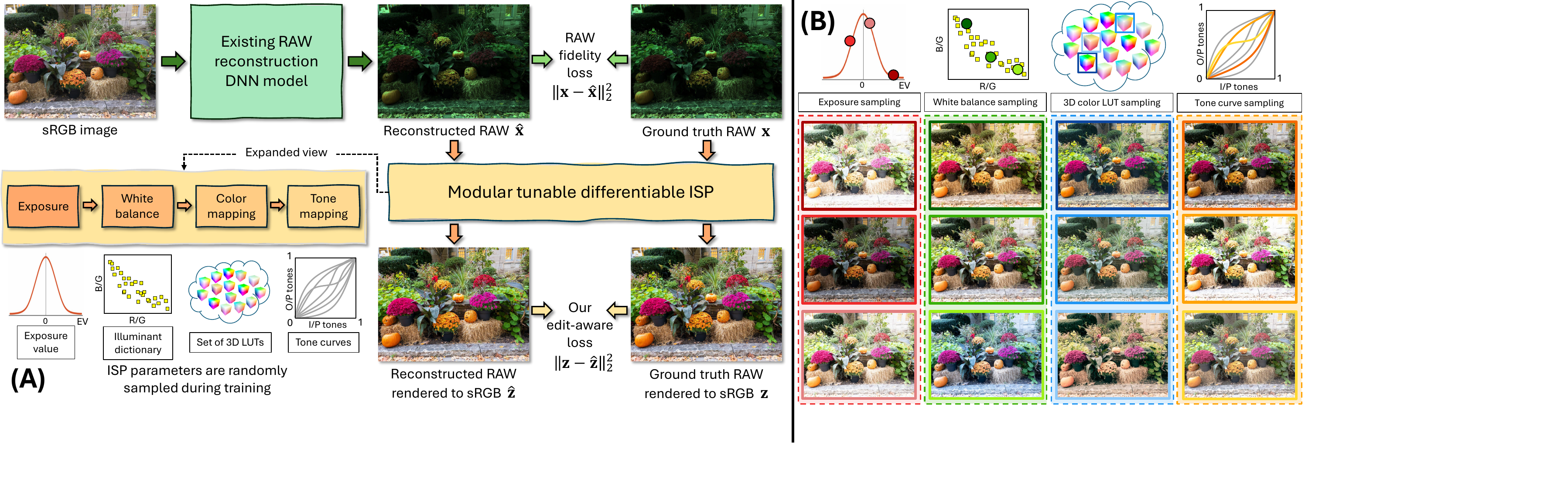}
  \caption{Overview of our edit-aware loss framework and differentiable ISP design. (A) Conventional RAW reconstruction models optimize for per-pixel RAW fidelity (green path), which often yields suboptimal results under diverse rendering or editing operations. We introduce an edit-aware loss (yellow path) that complements existing objectives by promoting robustness across different rendering styles. Our loss pipeline mimics a modular tunable differentiable ISP whose module parameters---exposure, white balance, color, and tone---are randomly sampled from distributions modeling realistic ISP settings. Both the ground-truth and reconstructed RAWs are rendered to sRGB using the same sampled parameters, and the loss is computed in sRGB space to encourage edit-consistent reconstruction. (B) Each column shows three samples illustrating the diversity of exposure, white balance, color, and tone adjustments encountered during training.  
  }\label{fig:method}
%  \vspace{-5mm}
\end{figure*}

We first review metadata-assisted approaches to RAW reconstruction, which leverage additional capture-time information to achieve higher accuracy than purely blind methods. One of the earliest works~\cite{yuan2011high} stored a downsampled RAW image and performed guided upsampling to recover the full-resolution RAW. Nguyen and Brown~\cite{nguyen2016raw,nguyen2018raw} computed and stored metadata in the form of estimated parameters modeling a global transformation from sRGB to RAW. Punnappurath and Brown~\cite{sam} instead saved a sparse set of uniformly sampled RAW values and used radial basis function interpolation to capture non-global ISP effects. Li et al.~\cite{li2023metadata} replaced this analytical interpolation with a neural implicit representation to better model nonlinear mappings. CAM~\cite{cam} improved upon fixed sampling by introducing content-aware sampling and a jointly trained decoder to enhance spatial adaptivity. Wang et al.~\cite{r2lcm,wang2024beyond} further proposed storing metadata as compact RAW latent features instead of direct pixel samples. Although these methods differ in sampling strategy and model class, they uniformly optimize for per-pixel RAW fidelity (sometimes with auxiliary regularizers, such as super-pixel losses for content-aware sampling~\cite{cam} or entropy coding for adaptive bit allocation~\cite{r2lcm}) and do not account for the need to re-render the reconstructed RAW to sRGB. In contrast, our edit-aware loss operates directly in sRGB space, optimizing reconstructed RAWs for rendering and editing robustness.

A complementary line of work explores blind RAW reconstruction methods that operate solely on the sRGB input~\cite{upi,nam2017modelling,ciexyznet,cycleisp,invisp,paramisp,conde2022model,rawdiff,conde2022reversed}. While RAW recovery targeting machine vision tasks, such as object detection, have been explored~\cite{reraw}, most methods focus exclusively on reconstruction accuracy without task-specific objectives. Early work on RAW reconstruction was calibration-based~\cite{debevec2008recovering,mitsunaga1999radiometric,grossberg2003determining,chakrabarti2014modeling,Chakrabarti2009empirical,kim2012new}. Later methods, such as UPI~\cite{upi}, reversed the sRGB image back to RAW by inverting the ISP stage-by-stage using non-learnable parametric operations. 
%Nam et al.~\cite{nam2017modelling} introduced one of the first deep models to learn the forward and inverse ISP transformations jointly. Subsequent methods such as CIE-XYZ-Net~\cite{ciexyznet}, CycleISP~\cite{cycleisp}, InvISP~\cite{invisp}, ParamISP~\cite{paramisp}, and model-based ISP approaches~\cite{conde2022model} extended this paradigm using convolutional and differentiable model-based architectures trained on paired RAW–sRGB datasets. 
Nam et al.~\cite{nam2017modelling} introduced a deep model trained on paired RAW-sRGB data for jointly learning forward and inverse ISP transformations, later extended by methods such as CIE-XYZ-Net~\cite{ciexyznet}, CycleISP~\cite{cycleisp}, InvISP~\cite{invisp}, ParamISP~\cite{paramisp}, and model-based ISP approaches~\cite{conde2022model}. Recently, diffusion-based methods~\cite{rawdiff} have also been explored for RAW reconstruction, but they continue to optimize losses in RAW space. While the forward-inverse approaches~\cite{nam2017modelling,ciexyznet,cycleisp,invisp,paramisp,conde2022model} sometimes include cyclic consistency loss terms to enforce sRGB reconstruction from the recovered RAW, the supervision always corresponds to the same input sRGB without any edits. Consequently, they cannot explicitly account for diverse photofinishing styles. Our work differs by introducing a stochastic, edit-aware loss computed in sRGB space via a tunable differentiable ISP, thereby encouraging robustness across diverse rendering styles and enabling accurate post-capture RAW editing targeting higher-quality sRGB outputs---the practical end goal for most RAW recovery workflows.

% Existing methods design their loss primarily to maximize pixel-level RAW reconstruction accuracy. Our approach departs from this objective by introducing a loss that optimizes for RAW reconstructions that enable accurate post-capture RAW editing targeting higher-quality sRGB outputs---the practical end goal for most RAW recovery workflows.
%-------------------------------------------------------------------------

\section{Methodology}
\label{sec:methodology}

We first briefly review the loss objectives used in existing RAW reconstruction methods. We then describe the ISP modules that form our edit-aware loss pipeline and explain how their tunable parameters are sampled during training.

\begin{figure*}
  \centering
  \includegraphics[width=0.95\textwidth]{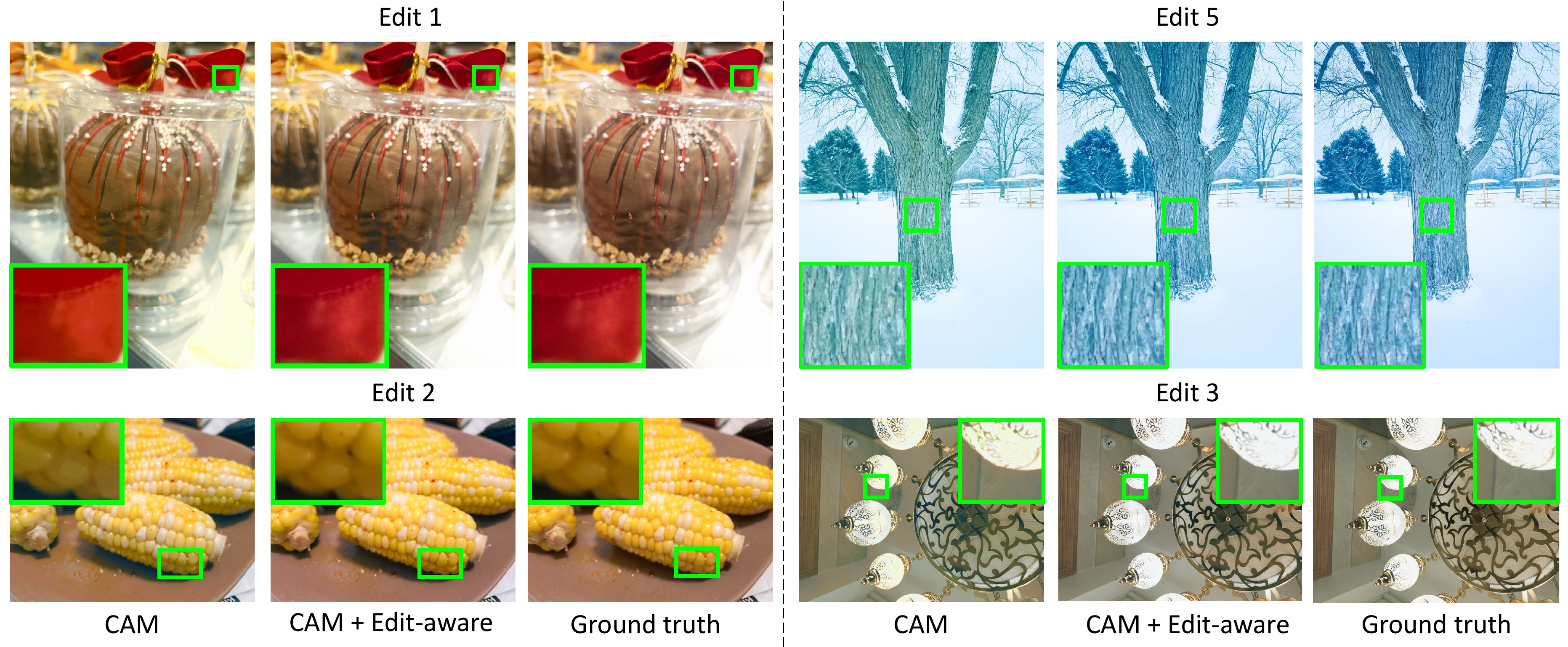}
  \caption{Results of adding our edit-aware loss to the RAW reconstruction method in CAM~\cite{cam}. See Table~\ref{tab:edits} for details of the edits. While the baseline model suffers from banding and color distortions, adding our edit-aware loss improves color and tone reproduction accuracy.}
  \label{fig:cam}
 \vspace{-2.5mm}
\end{figure*}

Let $\mathbf{x}$ denote the ground truth RAW image and $\mathbf{y}$ is the ISP's sRGB rendering of $\mathbf{x}$, then the objective of RAW reconstruction is to recover the RAW image $\hat{\mathbf{x}}$, given $\mathbf{y}$, as:
\begin{equation}
    \hat{\mathbf{x}}  = f_\theta(\mathbf{y}),
\end{equation}
where $f_\theta: \mathcal{Y} \to \mathcal{X}$ denotes the RAW reconstruction DNN model with learnable parameters $\theta$. In the case of metadata-assisted RAW reconstruction methods, the model $f_\theta$ may receive additional capture-time information as input, in addition to the sRGB image $\mathbf{y}$. Existing methods are trained to minimize the RAW reconstruction error between the ground truth RAW $\mathbf{x}$ and the network's prediction $\hat{\mathbf{x}}$:
\begin{equation}
\mathcal{L}_\mathrm{RAW}(\mathbf{x}, \hat{\mathbf{x}}) 
        = \| \mathbf{x} - \hat{\mathbf{x}} \|_2^2,
\end{equation}
typically, using $\ell_2$ error.

As shown in Fig.~\ref{fig:method}(A), we propose adding a loss function $\mathcal{L}_\mathrm{sRGB}$ computed between the ground truth RAW $\mathbf{x}$ and the network's prediction $\hat{\mathbf{x}}$, both rendered to sRGB using our differentiable ISP pipeline $g_\phi: \mathcal{X} \to \mathcal{Y}$.
\begin{equation}
\begin{split}
    \mathbf{z}  = g_\phi(\mathbf{x}), \quad \hat{\mathbf{z}}  = g_\phi(\hat{\mathbf{x}}), 
    \\
    \mathcal{L}_\mathrm{sRGB}(\mathbf{z}, \hat{\mathbf{z}}) 
        = \| \mathbf{z} - \hat{\mathbf{z}} \|_2^2.
\end{split}
\end{equation}
Here, the parameters $\phi$ of the ISP $g_\phi$ are not learned but sampled during training from distributions designed to closely mimic realistic ISP renderings such that the output of $g_\phi$ lies in the same distribution as $\mathbf{y}$ i.e., $\mathbf{z}, \hat{\mathbf{z}}, \mathbf{y} \in \mathcal{Y}$.

Our edit-aware loss pipeline $g_\phi$ models four ISP modules, namely (i) exposure $e_\varepsilon$, (ii) white balance $w_{\boldsymbol{\omega}}$, (iii) color manipulation $c_\rho$, and (iv) tone mapping $t_\tau$, such that: 
\begin{equation}
\phi = (\varepsilon, \boldsymbol{\omega}, \rho, \tau), \quad
g_\phi = t_\tau \circ c_\rho  \circ w_{\boldsymbol{\omega}} \circ e_\varepsilon. 
\end{equation}
Note that on a typical ISP, black-, white-level normalization, demosaicing and denoising are the first set of operations to be performed. Most RAW reconstruction methods recover a 3-channel denoised RAW image normalized to the range $[0,1]$. As such, we assume normalization, demosaicing and denoising have already been applied, and do not include these stages in our loss pipeline. Without loss of generality, in the subsequent discussion, we adopt a vectorized notation for all images $\in \mathbb{R}^{3 \times N}$, where $N$ denotes the total number of pixels in the image. The structure of our tunable and modular ISP framework and the details of these four modules are described next.

\begin{figure*}
  \centering
  \includegraphics[width=0.95\textwidth]{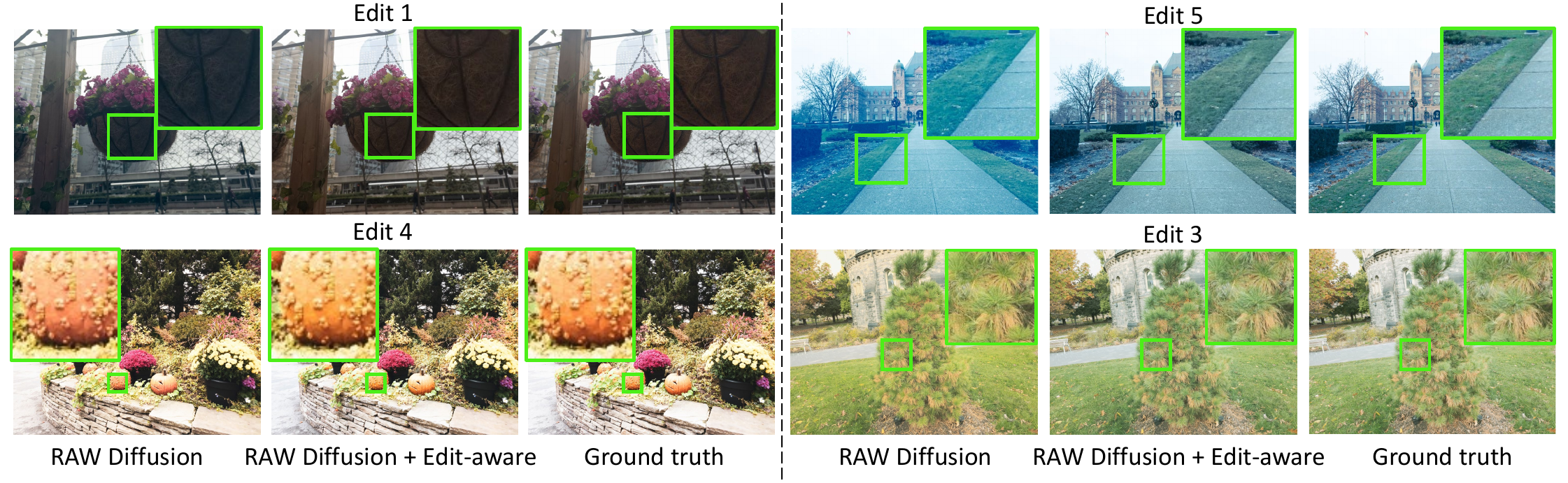}
  \caption{Results of adding our edit-aware loss to the RAW recovery method in RAW Diffusion~\cite{rawdiff}. See Table~\ref{tab:edits} for details of the edits. Compared to the baseline, which shows pronounced global color distortions, our edit-aware method more closely matches the ground truth.
  }\label{fig:rawdiff}
 \vspace{-2.5mm}
\end{figure*}

%---------------------------------------------------------------------

\subsection{Exposure module}
Exposure adjustments are typically applied directly to the RAW sensor image. We model exposure as:
\begin{equation}
    \mathbf{p_1} = e_\varepsilon(\mathbf{p_0}) = \mathbf{p_0} \cdot 2^{\varepsilon}, \ \  \varepsilon \sim \mathcal{N}(0, \sigma^2), \ \  \mathbf{p_0} \in \{\mathbf{x}, \hat{\mathbf{x}}\}, 
\end{equation}
where $\varepsilon$ represents the exposure value. This parameter is randomly sampled from a normal distribution $\mathcal{N}$ for each image in each mini-batch during training. Note that the same random sample is applied to $\mathbf{x}$ and $\hat{\mathbf{x}}$. Examples of exposure sampling are shown in the first column of Fig.~\ref{fig:method}(B).

\subsection{White-balance module}
Cameras apply a white-balance correction to the RAW image to remove the color cast due to the scene illumination. The idea is to divide each color channel of the image by the scene illumination, producing an image that appears lit under neutral illumination. However, users may still wish to adjust the white-point post-capture for aesthetic reasons (e.g., if they prefer a warmer or cooler image).

The illuminant estimated by the ISP is stored in the `AsShotNeutral' (ASN) tag in the RAW DNG file. To randomly sample an illuminant during training, we construct an illuminant dictionary $\mathcal{D}$ by extracting the ASN tags from a set of DNGs captured under different illuminations. We fit a 2D multivariate Gaussian distribution of joint chromaticity values $[ \frac{r}{g}, \frac{b}{g} ]$ to $\mathcal{D}$, and randomly sample an illuminant $\boldsymbol{\omega}$ from this distribution as follows:
\begin{equation}
\boldsymbol{\omega}
\thicksim
\mathcal{N} \left( \boldsymbol{\mathbf{\mu}}, \boldsymbol{\mathbf{\Sigma}} \right),
\end{equation}
\begin{equation}
\boldsymbol{\mathbf{\Sigma}} =\frac{1}{M}\sum_{m=1}^M\left( \left( \left[ \frac{r}{g},\frac{b}{g} \right]_m-\boldsymbol{\mathbf{\mu}}\right)^\intercal \left( \left[ \frac{r}{g},\frac{b}{g} \right]_m-\boldsymbol{\mathbf{\mu}} \right) \right),
\end{equation}
where $\boldsymbol{\mathbf{\mu}}$ and $\boldsymbol{\mathbf{\Sigma}}$ are the mean and covariance of the normalized chromaticity values in $\mathcal{D}$, $M$ is the number of illuminants in $\mathcal{D}$, and $\boldsymbol{\omega}, \boldsymbol{\mathbf{\mu}} \in \mathbb{R}^2$ and $\boldsymbol{\mathbf{\Sigma}}\in \mathbb{R}^{2\times 2}$.
We enforce that the sampled illuminant $\boldsymbol{\omega}$ falls within the convex hull of $\mathcal{D}$. We further constrain it to lie within a Euclidean threshold in chromaticity space relative to the ASN value of the image, because users typically make only small adjustments to the white balance during editing. White balance is performed by multiplying the image by a $3\!\times\!3$ diagonal matrix $W$ constructed directly from the sampled illuminant $\boldsymbol{\omega}$.

% \begin{figure*}
%   \centering
%   \includegraphics[width=\textwidth]{figs/ALL_unet_v3_400_rep0_v2.pdf}
%   \caption{Results of adding our edit-aware loss to a UNet-based~\cite{unet} RAW reconstruction method. See Table~\ref{tab:edits} for details of the edits.
%   }\label{fig:unet}
% %  \vspace{-5mm}
% \end{figure*}

Next, the white-balanced RAW image is transformed to the canonical CIE XYZ space using a full $3\!\times\!3$ matrix $C$ called the color space transform (CST) matrix. The CST matrix is dependent on the illumination $\boldsymbol{\omega}$ and is interpolated between two factory-calibrated color matrices specified in the DNG metadata~\cite{dng_spec}. We apply the matrices as:
% These operations are expressed as:
\begin{equation}
    \mathbf{p_2} = w_{\boldsymbol{\omega}}(\mathbf{p_1}) = C_{\boldsymbol{\omega}} W_{\boldsymbol{\omega}} \mathbf{p_1}.
\end{equation}
Examples of our white balance sampling are provided in the second column of Fig.~\ref{fig:method}(B).

\subsection{Color-manipulation module}
Color and tone manipulations are typically performed using lookup tables (LUTs) on conventional ISPs~\cite{hakki}. Specifically, color manipulations to impart a desired photofinishing style are implemented using 3D LUTs. A 3D LUT approximates a nonlinear 3D color transform by sparsely sampling the transformation on a discrete 3D lattice, and acts as a global color operator that maps a source RGB color to a target RGB color as $\tilde{c}([r,g,b]) = [r',g',b']$~\cite{zeng2020learning,lin2012nonuniform}. However, LUTs are not differentiable; therefore, we approximate the LUT using a multi-layer perceptron (MLP) $c$, similar to ~\cite{conde2024nilut}, such that $c$ is continuous and differentiable, and $c([r,g,b]) \approx \tilde{c}([r,g,b])$. We select a set of $K$ 3D LUTs and train an MLP corresponding to each to obtain $\{ {c_\rho} \}_{\rho=1}^K$. Note that the MLPs are learned offline and their weights are frozen during the training of the RAW reconstruction network. We pick a random LUT for each mini-batch and apply the color mapping as:
\begin{equation}
    \mathbf{p_3} = c_\rho(\mathbf{p_2}), \quad \rho \sim \mathcal{U}\{1, \dots, K\}, 
\end{equation}
where $\mathcal{U}$ denotes a uniform distribution. Representative examples are shown in the third column of Fig.~\ref{fig:method}(B).

\subsection{Tone-mapping module}
A tone curve is typically specified as a 1D LUT that is applied uniformly to all three color channels. We train an MLP $\psi$ that approximates the Adobe tone curve~\cite{hakki}, which we select as our default tone operator. We then introduce random perturbations around this tone curve by applying low-degree polynomials. Specifically, we generate random polynomials $S$ with degree $\tau$ that are monotonically non-decreasing. Finally, we apply a fixed $3\!\times\!3$ matrix $T$ to map from CIE XYZ space to linear-sRGB and a gamma function to obtain the final sRGB image.
\begin{equation}
    \mathbf{p_4} = t_\tau(\mathbf{p_3}) = (TS_\tau \psi(\mathbf{p_3}))^{\frac{1}{2.2}}, \ \  \tau \sim \mathcal{U}\{1, \dots, d\}. 
\end{equation}
Examples of tone sampling can be seen in the last column of Fig.~\ref{fig:method}(B). Although we do not explicitly model local tone mapping for the sake of simplicity and speed, most RAW reconstruction methods adopt patch-based training, and our random sampling of the tone curves for individual patches can offer a degree of robustness to local tonal adjustments. 

\subsection{Combined loss}
The combined loss is computed as:
\begin{equation}
\mathcal{L}_\mathrm{total} = \mathcal{L}_\mathrm{RAW}(\mathbf{x}, \hat{\mathbf{x}}) 
        + \mathcal{L}_\mathrm{misc} + \lambda \mathcal{L}_\mathrm{sRGB}(\mathbf{z}, \hat{\mathbf{z}}),
\label{eqn:final_loss}
\end{equation}
where $\lambda$ is a scalar weighting factor, and $\mathcal{L}_\mathrm{misc}$ denotes any other additional losses employed by the selected RAW recovery method that do not directly target per-pixel RAW image reconstruction. For example, CAM\cite{cam} uses a super-pixel loss for content-aware sampling. 

Our four tunable ISP modules provide a lightweight yet expressive model of real-world ISP variability, enabling our loss $\mathcal{L}_\mathrm{sRGB}$ to deliver edit-aware supervision that generalizes across diverse rendering styles and editing workflows.
%Coupled with our four tunable ISP modules, which provide a lightweight yet expressive model of real-world ISP variability, our loss $\mathcal{L}_\mathrm{sRGB}(\mathbf{z}, \hat{\mathbf{z}})$ delivers edit-aware supervision that generalizes across rendering styles and editing workflows.

%-------------------------------------------------------------------------

\section{Results}
\label{sec:results}

We assess the effectiveness of our edit-aware loss on two representative RAW reconstruction methods from prior work---CAM~\cite{cam} and RAW Diffusion~\cite{rawdiff}---to demonstrate its generality and impact. CAM serves as a metadata-assisted approach, while RAW Diffusion represents a blind model. Additionally, we designed a vanilla UNet–based~\cite{unet} model to further evaluate our loss in a metadata-assisted setting. 

We use the smartphone RAW dataset from~\cite{s24} for training and evaluation. The dataset contains 3,224 images captured with the main rear camera of a Samsung S24 Ultra, encompassing a diverse range of scenes and lighting conditions. Each image has a resolution of $3000\!\times\!4000$ pixels. For training pairs, we use the provided denoised and demosaiced linear RAW images and the corresponding camera-ISP-rendered sRGB JPEGs. The dataset is split into 2,619 training, 205 validation, and 400 test images.

We fix the parameters defining the sampling distributions of our modular ISP across all experiments. Specifically, we set $\sigma = 0.75$,  $M=2,619$ (corresponding to all training images), $K=15$, and $d=5$. 
For the approximator MLPs, we follow the architecture and training protocol of~\cite{conde2024nilut}: the LUT MLP $c$ uses three input and output features, and the tone MLP $\psi$ uses one input and output feature.
The edit-aware loss $\mathcal{L}_\mathrm{sRGB}$ is computed using the standard $\ell_2$ error.

For evaluation, reconstructed RAW images are saved in DNG format~\cite{dng} and edited using Adobe Photoshop.
Photoshop represents a full-featured, complex software ISP that is independent of our training pipeline. Our method does not assume access to or differentiability of the ISP used at inference. We evaluate using Photoshop to demonstrate robustness to unknown processing.
%While our method employs a simplified ISP, Photoshop represents a full-featured, complex software ISP; nonetheless, our approach demonstrates strong performance as we shall show in our experiments. 
We select edits that are diverse, realistic, and representative of common user adjustments. Specifically, we apply the five edits listed in Table~\ref{tab:edits} using Adobe Camera RAW, Photoshop’s plugin for RAW image editing. Each edit specifies the adjusted settings or parameters, and their values; a single edit may modify multiple parameters simultaneously. For example, `Presets' in Table~\ref{tab:edits} adjust several ISP parameters at once to achieve a particular style. For consistency and reproducibility, the same fixed set of edits are applied to all test images.

% (1) Edit 1: Adobe Camera RAW default rendering, (2) Edit 2: Preset $\to$ Color $\to$ Bright, (3) Edit 3: Preset $\to$ Creative $\to$ Flat and green, Light $\to$ Exposure $+0.5$, (4) Edit 4: Preset $\to$ Creative $\to$ Warm contrast, Light $\to$ Exposure $-0.25$, Curve $\to$ Lights $+50$, (5) Edit 5: Preset $\to$ Creative $\to$ Cool matte, Light $\to$ Exposure $+1.5$, Curve $\to$ Darks $+50$, Color $\to$ Temperature $3500$.
\begin{table}[]
\centering
\resizebox{0.85\columnwidth}{!}{%
\begin{tabular}{l|l}
\toprule
Name & Adobe Camera RAW settings \\
\midrule
Edit 1 & Adobe Camera RAW default rendering \\
\midrule
Edit 2 & Preset $\to$ Color $\to$ Bright \\
\midrule
\multirow{2}{*}{Edit 3}  & Preset $\to$ Creative $\to$ Flat and green \\
       & Light $\to$ Exposure $\to$ $+0.5$ \\
\midrule
\multirow{3}{*}{Edit 4} & Preset $\to$ Creative $\to$ Warm contrast \\
       & Light $\to$ Exposure $\to$ $-0.25$ \\
       & Curve $\to$ Lights $\to$ $+50$ \\
\midrule
\multirow{4}{*}{Edit 5} & Preset $\to$ Creative $\to$ Cool matte \\
       & Light $\to$ Exposure $\to$ $+1.5$ \\
       & Curve $\to$ Darks $\to$ $+50$ \\
       & Color $\to$ Temperature $\to$ $3500$K \\
\bottomrule
\end{tabular}
}
\caption{A description of the five different edits applied at inference to the RAW DNGs. The settings are applied in Adobe Camera RAW, which is Adobe's plugin for RAW image processing. \label{tab:edits}
}
\end{table}

\begin{table*}[]
\centering
\resizebox{\textwidth}{!}{%
\begin{tabular}{l|l|lll|lll|lll|lll|lll}
\toprule
\multirow{2}{*}{Method}         & RAW       & \multicolumn{3}{c|}{Edit 1} & \multicolumn{3}{c|}{Edit 2} & \multicolumn{3}{c|}{Edit 3} & \multicolumn{3}{c|}{Edit 4} & \multicolumn{3}{c}{Edit 5} \\
               & PSNR & PSNR   & SSIM   & $\Delta$E   & PSNR  & SSIM   & $\Delta$E   & PSNR   & SSIM   & $\Delta$E   & PSNR   & SSIM   & $\Delta$E   & PSNR   & SSIM   & $\Delta$E   \\
\toprule
CAM            &   37.17          &             27.27 & 0.8644 & 7.12 & 26.95 & 0.8632 & 7.48 & 28.91 & 0.8951 & 5.91 & 27.70 & 0.8999 & 6.12 & 25.43 & 0.8456 & 8.00      \\
CAM + Edit-aware     &    \yb{37.57}       &             \yb{29.24} & \yb{0.8860} & \yb{5.20} & \yb{28.97} & \yb{0.8819} & \yb{5.54} & \yb{30.72} & \yb{0.9068} & \yb{4.33} & \yb{29.23} & \yb{0.9121} & \yb{4.54} & \yb{27.43} & \yb{0.8637} & \yb{5.98}    \\
\midrule
RAWDiff        &    \yb{34.18}        &             24.27 & 0.8584 & 9.50 & 23.96 & 0.8608 & 9.99 & 26.09 & 0.8965 & 8.25 & 25.33 & \yb{0.9037} & 8.21 & 23.29 & 0.8484 & 9.91     \\
RAWDiff + Edit-aware &     33.37       &              \yb{25.44} & \yb{0.8728} & \yb{8.52} & \yb{25.46} & \yb{0.8730} & \yb{8.74} & \yb{27.17} & \yb{0.9014} & \yb{7.26} & \yb{25.66} & 0.9019 & \yb{7.74} & \yb{25.03} & \yb{0.8590} & \yb{8.60}     \\
\midrule
UNet           &     \yb{38.82}      &             28.52 & 0.8693 & 5.89 & 28.17 & 0.8664 & 6.38 & 30.09 & 0.8956 & 4.90 & 28.73 & 0.9016 & 4.93 & 26.44 & 0.8463 & 6.88     \\
UNet + Edit-aware    &    35.62       &             \yb{29.26} & \yb{0.8832} & \yb{5.30} & \yb{29.01} & \yb{0.8793} & \yb{5.68} & \yb{30.83} & \yb{0.9032} & \yb{4.36} & \yb{29.32} & \yb{0.9097} & \yb{4.48} & \yb{28.02} & \yb{0.8635} & \yb{5.75}    \\
\bottomrule
\end{tabular}
}
\caption{Quantitative evaluation of the proposed edit-aware loss integrated into different RAW reconstruction frameworks.
Results are reported for CAM~\cite{cam}, RAW Diffusion~\cite{rawdiff}, and a UNet–based~\cite{unet} model, evaluated on 400 test images from the RAW smartphone dataset of~\cite{s24}. We report PSNR (dB) for reconstructed RAWs, and PSNR (dB), SSIM, and $\Delta$E (lower is better) for the corresponding edited sRGB renderings. Editing operations follow the settings described in Table~\ref{tab:edits}. The \colorbox{yellow!35}{\textbf{best}} results are highlighted.
\label{tab:main}
}
\end{table*}

\begin{table*}[]
\centering
\resizebox{0.775\textwidth}{!}{%
\begin{tabular}{l|lll|lll|lll}
\toprule
\multirow{2}{*}{Method}      & \multicolumn{3}{c|}{EV +2} & \multicolumn{3}{c|}{CCT 3000K} & \multicolumn{3}{c}{EV +2 \& CCT 3000K} \\
           & PSNR   & SSIM   & $\Delta$E   & PSNR    & SSIM   & $\Delta$E   & PSNR    & SSIM   & $\Delta$E   \\
\midrule
UNet: FT image       &             30.26 & 0.8777 & 3.39 & 30.84 & 0.8859 & 3.79 & 30.12 & 0.8756 & 3.44    \\

UNet + Edit-aware: FT image &            \rb{31.31} & \rb{0.8889} & \rb{3.17} & \rb{31.11} & \yb{0.8898} & \rb{3.73} & \rb{31.09} & \rb{0.8861} & \rb{3.23}    \\
UNet + Edit-aware: FT image \& FT edit     &   \yb{31.51} & \yb{0.8894} & \yb{3.04}  & \yb{31.20} & \yb{0.8898} & \yb{3.71}      & \yb{31.26} & \yb{0.8870} & \yb{3.18}  \\
\bottomrule
\end{tabular}
}
\caption{Quantitative results of fine-tuning (FT) a UNet-based~\cite{unet} metadata-assisted RAW reconstruction model. Results are averaged over the 400 test images from the RAW smartphone dataset of~\cite{s24}. Metrics include PSNR (dB), SSIM, and $\Delta$E for the edited sRGB renderings. EV denotes exposure value and CCT represents correlated color temperature. The \colorbox{yellow!35}{\textbf{best}} and \colorbox{red!15}{second-best} results are highlighted.
\label{tab:finetuning}
}
\end{table*}

We first evaluate our edit-aware loss using the metadata-assisted CAM~\cite{cam} method. CAM employs a learned sampler at capture time to select a subset of RAW pixels. These RAW samples, saved as metadata, guide the decoder at inference to reconstruct the full RAW image from the sRGB input. Following the training procedure and hyperparameters recommended by the authors, we train two separate models: a baseline model using the original CAM implementation and an edit-aware variant under identical settings, augmented with our loss (we set $\lambda = 2$ in Eq.~\ref{eqn:final_loss}). Evaluation is performed on the 400-image test split of the smartphone RAW dataset~\cite{s24}. In Table~\ref{tab:main}, we report RAW reconstruction quality using PSNR (dB) and evaluate the sRGB renderings after Photoshop edits using PSNR, SSIM~\cite{ssim}, and $\Delta$E~\cite{deltae} (lower is better). Our method consistently improves all metrics across edits, with the largest gains---up to 2 dB---observed under more substantial edits, such as Edit 5. Interestingly, RAW PSNR also shows a slight improvement. Qualitative results are shown in Fig.~\ref{fig:cam}, where the baseline CAM model exhibits banding artifacts and color distortions due to collapsed tones, whereas our method produces smoother, more accurate color and tonal responses. See Section~\ref{sec:nus} of supplementary material for additional evaluations of the CAM model on the NUS dataset~\cite{NUS}.

Next, we perform a similar experiment using the RAW Diffusion~\cite{rawdiff} method, a blind reconstruction model that takes only the sRGB image as input. RAW Diffusion employs a diffusion-based framework with RGB-guided residual blocks to reconstruct high-fidelity RAW images. As before, we follow the training procedure and hyperparameters recommended by the authors, using a $\lambda = 4$ weighting factor for our edit-aware loss. Results are summarized in Table~\ref{tab:main}. Our method trades off a small amount of RAW fidelity but significantly improves sRGB reconstruction across different edits. Note that because RAW Diffusion is blind, overall performance is generally lower than metadata-assisted CAM, as expected. Qualitative results in Fig.~\ref{fig:rawdiff} show that our method achieves more accurate color reproduction and closely matches the ground truth, while the baseline exhibits noticeable global color deviations.

Finally, we evaluate our method in a metadata-assisted setting using a standard UNet~\cite{unet} architecture. The extra capture-time metadata is a downsampled version of the RAW image $\mathbf{x_d}$, and reconstruction is performed given the sRGB image and downsampled RAW: $\hat{\mathbf{x}} = f_\theta(\mathbf{y}, \mathbf{x_d})$. We use an 8$\times$ downsampling factor along both axes, corresponding to $\approx 1.5\%$ metadata overhead in RAW samples. The UNet receives a 6-channel input: the first three channels are the sRGB image, and the last three are the downsampled RAW image bilinearly upsampled back to the original image resolution. The network outputs the reconstructed RAW image. 
% Training details are provided in the supplementary.
% Move to supplementary:
% Training uses the Adam optimizer~\cite{Adam} with a learning rate of 0.001, a batch size of 32, and a patch size of 304. The model is trained for 50 epochs. There are 32 initial filters in the UNet. 
The baseline model is trained using the RAW reconstruction loss $\mathcal{L}_\mathrm{RAW}$, while for our method, we use only the edit-aware sRGB loss $\mathcal{L}_\mathrm{sRGB}$, omitting $\mathcal{L}_\mathrm{RAW}$ as a test of whether our loss in isolation can guide edit-robust RAW reconstruction. Results in Table~\ref{tab:main} show a tradeoff in RAW fidelity, as expected without direct RAW supervision, but our method produces significantly better reconstructions across edits. See Section~\ref{sec:additional_qualitative_results} of supplementary material for training details and qualitative results.
% Qualitative results in Fig.~\ref{fig:unet} illustrate color deviations (first row) and banding artifacts (last row) in the baseline model, whereas our results are closer to the ground truth.

\begin{figure}
  \centering
  \includegraphics[width=\columnwidth]{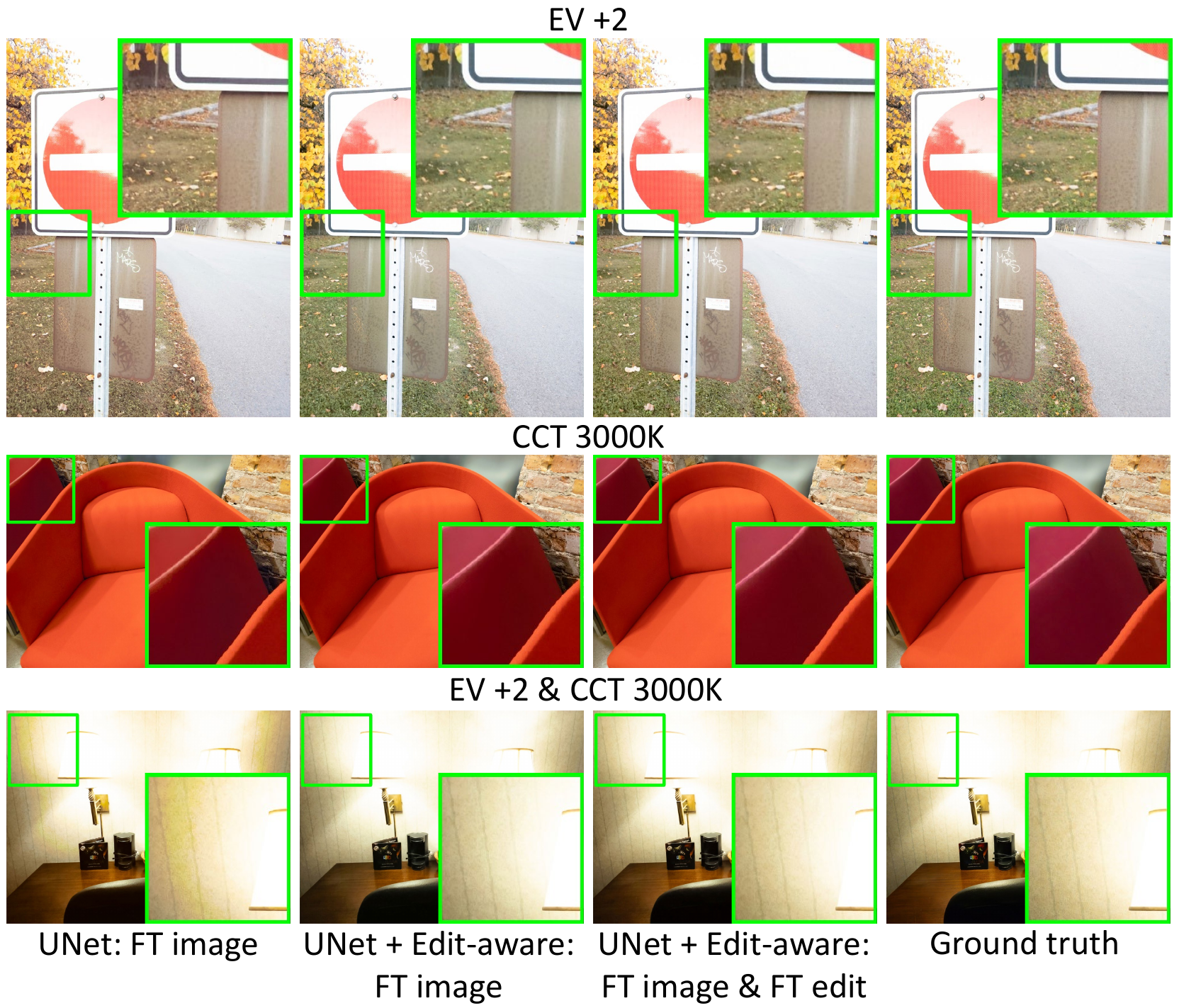}
  \caption{Results of fine-tuning (FT) a UNet-based~\cite{unet} metadata-assisted RAW reconstruction model using our edit-aware loss. Our loss enables fine-tuning not only on the specific test image but also with respect to the target edit, yielding reconstructions that better align with the desired photofinishing outcome.
  }\label{fig:finetuning}
 \vspace{-5mm}
\end{figure}

\subsection{Target edit-aware fine-tuning}
\label{sec:finetuning_unet}
Metadata-assisted methods offer the additional capability of model fine-tuning at inference time, typically using RAW samples stored as metadata, to adapt the model weights to a specific image. With our edit-aware loss, fine-tuning can be directed not only to the image, but also to the target edit. We demonstrate this ability of our method using the UNet model, where the fine-tuning loss is computed as $\mathcal{L}_\mathrm{sRGB-FT}(\mathbf{z_d}, \mathbf{\hat{z}_d}) = \| \mathbf{z_d} - \mathbf{\hat{z}_d} \|_2^2$, where $\mathbf{z_d}  = g_\phi(\mathbf{x_d})$, $\mathbf{\hat{z}_d}  = g_\phi(\mathbf{\hat{x}_d})$, and $\mathbf{\hat{x}_d}$ is obtained by downsampling the output of the network $\hat{\mathbf{x}}$. Importantly, during fine-tuning, the parameters $\phi$ can be fixed to match the target edit (e.g., $\varepsilon = 0.5$ for a +0.5 exposure edit) or randomly sampled as during training. We perform fine-tuning for 100 iterations using random 1024-pixel crops; $\mathbf{z_d}$ is computed once.
% and a reduced learning rate of 0.0001

We demonstrate this idea on exposure and white-balance edits, as the two most common and representative user adjustments. Table~\ref{tab:finetuning} shows three scenarios: the first row is the baseline model fine-tuned for the test image, the second row is our model fine-tuned with $\phi$ randomly sampled, and the last row is our model fine-tuned with $\phi$ fixed to the target edit. Even with standard fine-tuning (first and second rows), our method maintains an advantage over the baseline model, and fixing $\phi$ to the target edit (last row) further improves performance. The qualitative results in Fig.~\ref{fig:finetuning} show that target edit-aware fine-tuning produces more accurate and visually consistent outputs under the selected edits.

\begin{table}[]
\centering
\resizebox{0.8\columnwidth}{!}{%
\begin{tabular}{lccccc}
\multicolumn{6}{c}{Sampling \checkmarkk \quad \quad No sampling \circc \quad \quad Disabled \checkcross} \\
\toprule
\multirow{2}{*}{Model} & \multicolumn{4}{c}{Loss modules}         & sRGB \\

         & Exp. & WB & Color & Tone &  PSNR \\
\midrule
Exp. only  &    \checkmarkk      & \checkcross    &  \checkcross    & \checkcross    &       23.22    \\
WB only        & \checkcross         &   \checkmarkk & \checkcross     & \checkcross    &  22.35         \\
Color only      &   \checkcross       & \checkcross   & \checkmarkk     & \checkcross    &       20.54    \\
Tone only       &   \checkcross       &  \checkcross  &  \checkcross    & \checkmarkk    &        23.77    \\
Fixed pipeline &  \circc        &   \circc & \circc     &  \circc   &       24.20    \\
\midrule
\textbf{Ours}           &  \checkmarkk        & \checkmarkk   &  \checkmarkk    & \checkmarkk    &        \textbf{25.15}  \\
\bottomrule
\end{tabular}
}
\caption{
Ablation study on the contribution of different loss modules and sampling configuration. Results are reported using the UNet reconstruction model on a subset of 50 challenging images from the test split of the RAW smartphone dataset of~\cite{s24}. We report sRGB PSNR (dB) for Edit 5 (see Table~\ref{tab:edits} for edit details). \label{tab:ablation}
}
\vspace{-5mm}
\end{table}

\subsection{Ablations}
\label{subsec:ablations}
We perform ablations on the individual loss modules and the ISP parameter sampling to assess their contributions. Results are summarized in Table~\ref{tab:ablation} using the UNet model, evaluated on a subset of 50 challenging images from the test split. We report sRGB PSNR for Edit 5 (see Table~\ref{tab:edits}). The first four rows examine the contribution of each loss module. The fifth row tests a configuration using all modules but with parameter sampling disabled (i.e., a fixed ISP pipeline). This is equivalent to the cyclic consistency loss term used by forward-inverse methods~\cite{ciexyznet,cycleisp,invisp} that enforces sRGB reconstruction from the recovered RAW {\it without} any edit sampling. As shown in the final row, our complete configuration, including all modules with randomized parameter sampling, yields the best performance.

\subsection{User study}
We conducted a blind, forced-choice perceptual study in which participants selected between the results with and without our edit loss that best matched the edited ground truth in terms of color and brightness while minimizing artifacts. The study used 20 images from the challenging 50-image set in Section~\ref{subsec:ablations}. Each user rated all 20 images for the 3 methods in Table~\ref{tab:main}, with a random edit from Table~\ref{tab:edits} selected per image. We surveyed 25 users, yielding 1500 responses. Our method was preferred 83\% of the time; preferences for CAM~\cite{cam}, RAW Diffusion~\cite{rawdiff}, and UNet~\cite{unet} were 77\%, 88\%, and 82\%.

\subsection{Image-specific edits}
In Table~\ref{tab:edits}, we used a fixed edit set for consistency and batch efficiency in Photoshop. Practical edits are typically image- and content-specific. We additionally conducted an experiment where manual image-specific edits were applied to the difficult 50-image set in Section~\ref{subsec:ablations}. PSNRs reported in Table~\ref{tab:manual_edits} are lower due to more challenging images and edits but we observe the same improvement trend as with fixed edits, indicating good generalization to image-specific edits.
\begin{table}[!t]
%\setlength{\tabcolsep}{3pt}
% \vspace*{-0.65cm}
\centering
\resizebox{0.95\columnwidth}{!}{%
\begin{tabular}{l|ll|ll|ll}
\toprule
\multirow{2}{*}{Method}  & \multicolumn{2}{c|}{CAM~\cite{cam}}       & \multicolumn{2}{c|}{RAWDiff~\cite{rawdiff}} & \multicolumn{2}{c}{UNet~\cite{unet}}  \\
              & PSNR   & SSIM    & PSNR   & SSIM   &  PSNR  & SSIM       \\
\toprule
Model & 22.11 & 0.6499 & 20.57 & 0.6927 & 23.30 & 0.6794   \\
Model + EA & \yb{24.36} & \yb{0.7087}  & \yb{21.59} & \yb{0.7019}  & \yb{24.58} & \yb{0.7199}     \\
\bottomrule
\end{tabular}
}
\caption{
Results with manual image-specific edits applied to a subset of 50 challenging images from the RAW smartphone dataset of~\cite{s24}. Our edit-aware method (EA) generalizes well to image-specific edits across various RAW reconstruction methods.
\label{tab:manual_edits}
}
% \vspace*{-0.5cm}
\end{table}

%-------------------------------------------------------------------------

\section{Discussion and conclusion}
\label{sec:conclusion}

We presented an edit-aware loss framework that enhances the robustness of RAW reconstruction models under diverse rendering styles and post-capture edits. Unlike prior approaches that optimize solely for pixel-level RAW fidelity, our method explicitly aligns the training objective with downstream photographic editing by introducing a differentiable, modular, and tunable ISP that models realistic photofinishing variations. This plug-and-play loss formulation can be integrated into existing RAW reconstruction pipelines, improving edit fidelity and rendering consistency without any architectural changes. 

While the inclusion of our edit-aware loss increases training time due to the added ISP sampling (by approximately 15\% in our experiments), inference remains unchanged, allowing existing RAW reconstruction models to benefit from enhanced editability without impacting deployment efficiency. Although our framework introduces multiple ISP modules and diverse parameter sampling to cover a wide range of potential edits, the design remains highly flexible. For example, if specific target edit conditions are known {\it a priori}, the sampling can be restricted, enabling training for those specific types of edits while still leveraging the edit-aware supervision paradigm.

Extensive experiments demonstrate that our approach yields consistent gains across both metadata-assisted and blind reconstruction models, producing higher-quality sRGB renderings under a wide range of edits. Our method performs best for strong edits inducing large tonal, color, or contrast changes. Observed failures primarily occurred in extreme low-light scenes or uncommon lighting (e.g., green LEDs) not seen during training.
% Looking ahead, we envision that combining edit-aware objectives with future learned ISPs and generative reconstruction frameworks will further narrow the gap between computational photography and human-centric image editing.

\appendix
\newcommand{\hbAppendixPrefix}{S}
\renewcommand{\thefigure}{\hbAppendixPrefix\arabic{figure}}
\setcounter{figure}{0}
\renewcommand{\thetable}{\hbAppendixPrefix\arabic{table}}
\setcounter{table}{0}
\renewcommand{\theequation}{\hbAppendixPrefix\arabic{equation}}
\setcounter{equation}{0}
\renewcommand{\thesection}{\hbAppendixPrefix\arabic{section}}
\setcounter{section}{0}

\clearpage

\twocolumn[{%
\centering
\Large \textbf{Supplementary Material}\\[1.5em]
}]

%\maketitle

% \title{Supplementary Material \\ Edit-aware RAW Reconstruction}

% \maketitle

%-------------------------------------------------------------------------

\begin{figure*}
  \centering
  \includegraphics[width=0.95\textwidth]{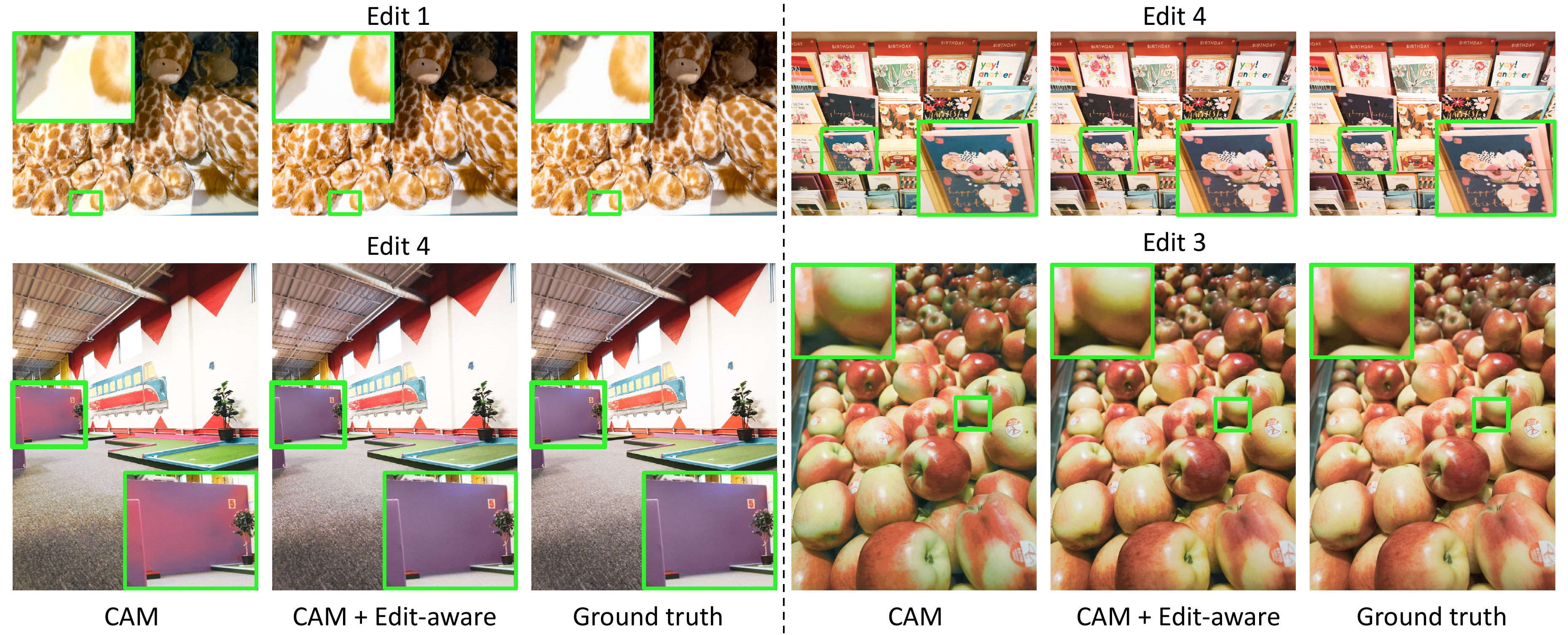}
  \caption{Additional qualitative results of adding our edit-aware loss to the RAW reconstruction method in CAM~\cite{cam}. This figure complements Fig.~\ref{fig:cam} of the main paper. See Table~\ref{tab:edits} of the main paper for details of the edits. The baseline method exhibits noticeable color shifts, whereas our approach produces smoother tones and more accurate colors that better match the ground truth. 
}
  \label{fig:cam_supp}
%  \vspace{-5mm}
\end{figure*}
% While the baseline model suffers from banding and color distortions, adding our edit-aware loss improves color and tone reproduction accuracy.

\begin{figure*}
  \centering
  \includegraphics[width=\textwidth]{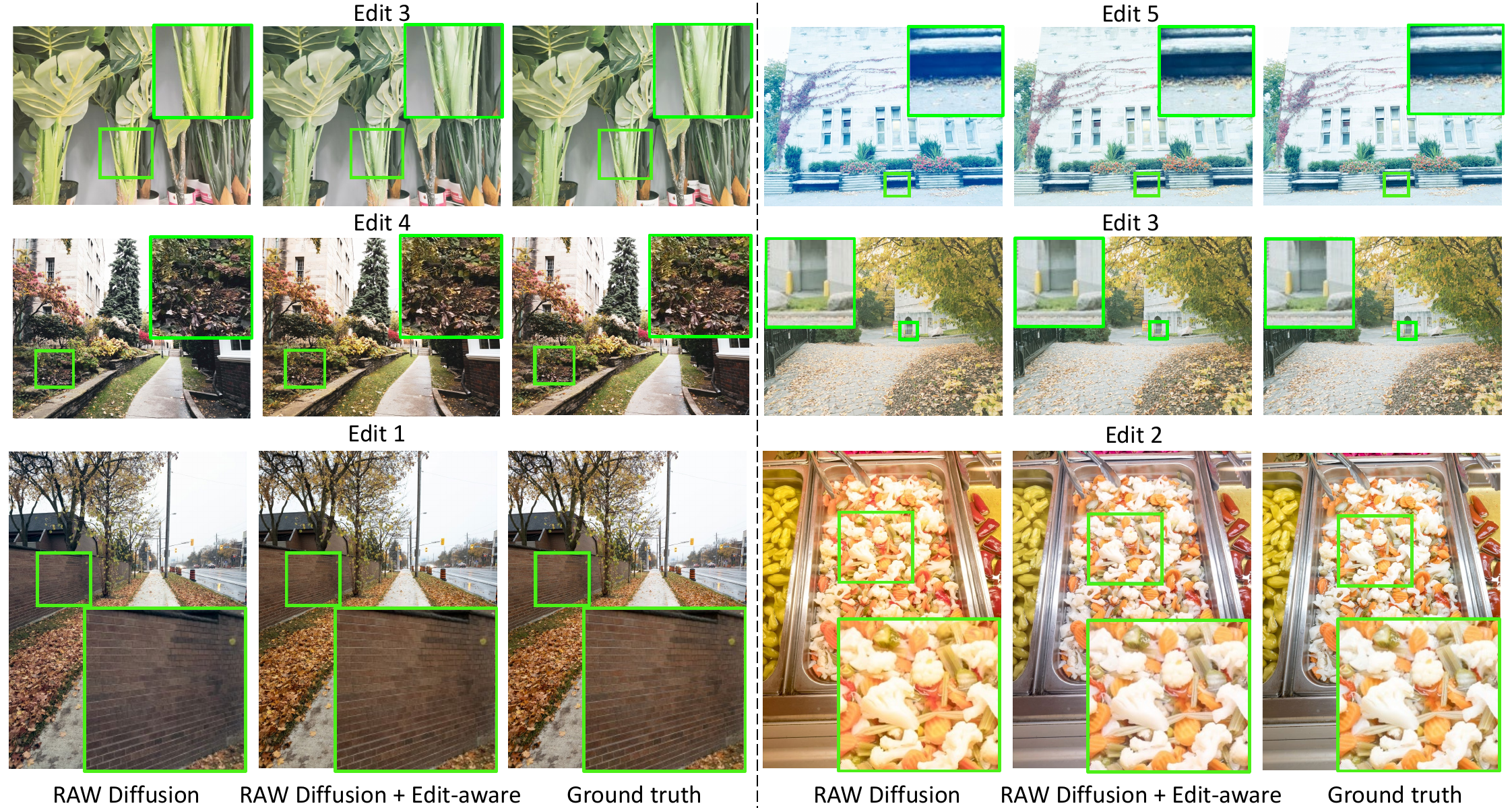}
  \caption{Additional qualitative results of adding our edit-aware loss to the RAW recovery method in RAW Diffusion~\cite{rawdiff}. These examples expand upon Fig.~\ref{fig:rawdiff} of our main paper. See Table~\ref{tab:edits} of the main paper for details of the edits. The baseline model often shows inconsistent or shifted colors, whereas our method yields more faithful color reproduction and balanced tones that better align with the ground truth.
  }\label{fig:rawdiff_supp}
%  \vspace{-5mm}
\end{figure*}

% Compared to the baseline, which shows pronounced global color distortions, our edit-aware method more closely matches the ground truth.

\begin{figure*}
  \centering
  \includegraphics[width=\textwidth]{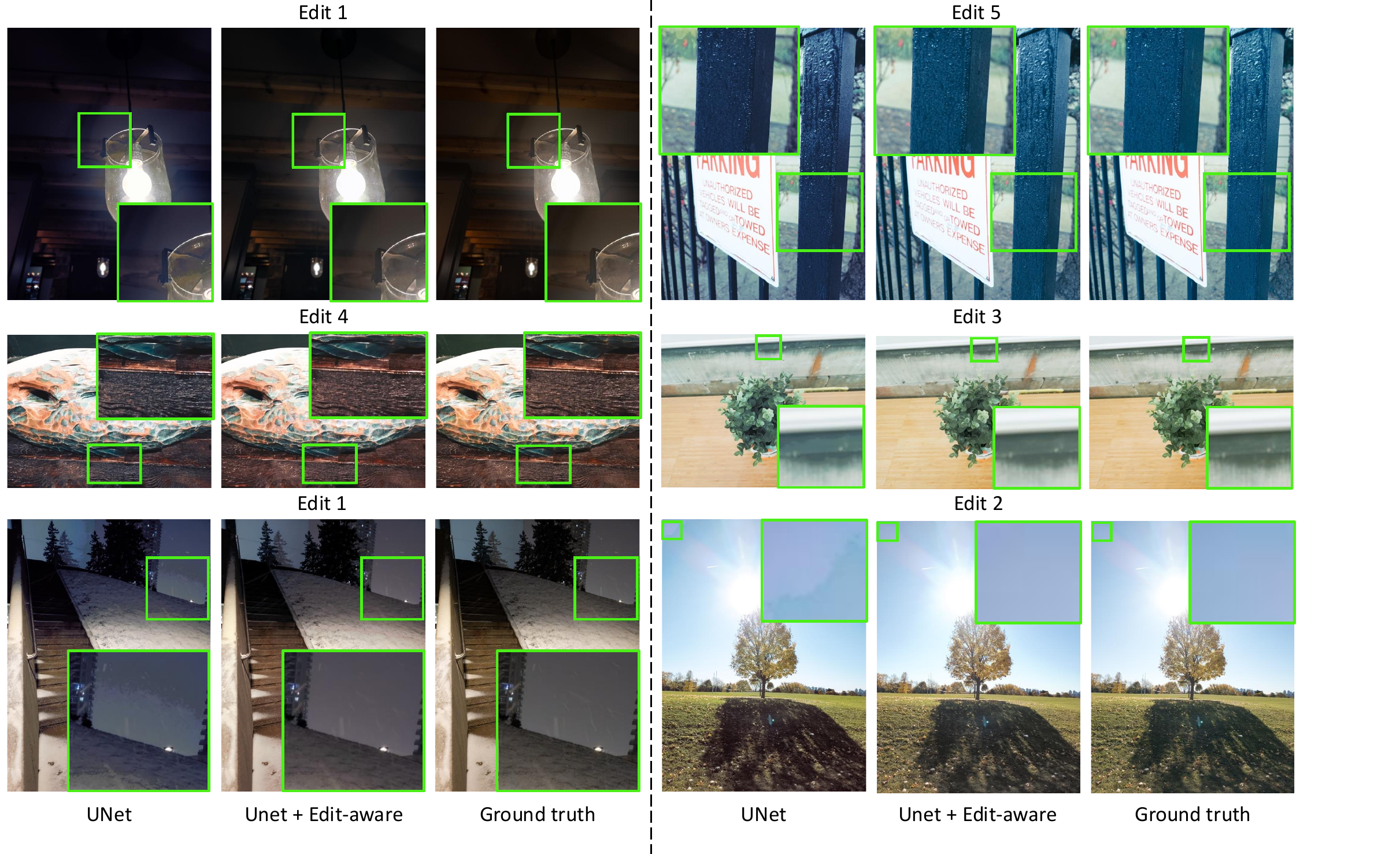}
  \caption{Results of adding our edit-aware loss to a UNet-based~\cite{unet} RAW reconstruction method. See Table~\ref{tab:edits} of the main paper for details of the edits. Compared to the baseline, which exhibits banding artifacts and shifted colors due to collapsed tones, our method produces smooth tonal transitions without quantization artifacts and more accurate, faithful colors in the rendered sRGB outputs.
  }\label{fig:unet_supp}
%  \vspace{-5mm}
\end{figure*}

This supplementary material provides additional experiments and analyses that complement the main paper. In Section~\ref{sec:additional_qualitative_results}, we show more qualitative examples of adding our edit-aware loss to different RAW reconstruction frameworks and provide implementation details for the UNet-based model. Section~\ref{sec:finetuning_cam} explores target-edit-aware fine-tuning using the CAM~\cite{cam} method. In Section~\ref{sec:other_edits}, we evaluate our approach on more diverse post-processing edits, such as dehazing and local tone mapping. Section~\ref{sec:nus} shows results on the NUS~\cite{NUS} dataset. Finally, Section~\ref{sec:additional_ablations} presents further ablation studies. Collectively, these results offer deeper insights into the effectiveness and versatility of our edit-aware loss.

\section{Additional results}
\label{sec:additional_qualitative_results}

Figs.~\ref{fig:cam_supp} and~\ref{fig:rawdiff_supp} provide additional results of our edit-aware loss integrated into the RAW recovery methods in CAM~\cite{cam} and RAW Diffusion~\cite{rawdiff}. These figures extend Figs.~\ref{fig:cam} and~\ref{fig:rawdiff} of our main paper. Across these examples, our method consistently reduces color and tone discrepancies, producing outputs that better reflect the applied edits.

Fig.~\ref{fig:unet_supp} shows similar experiments on the metadata-assisted UNet~\cite{unet} model. Compared to the baseline UNet, which has banding artifacts and color shifts, incorporating our edit-aware loss improves both tonal consistency and color fidelity in the rendered sRGB outputs. The UNet was trained using the Adam optimizer~\cite{Adam} with a learning rate of 0.001, batch size of 32, and patch size of 304. Training ran for 50 epochs. The fine-tuning experiments in Section~\ref{sec:finetuning_unet} of the main paper were conducted at a reduced learning rate of 0.0001. Following the original UNet design, the first convolutional block used 32 filters.

\section{Target-edit-aware fine-tuning on CAM~\cite{cam}}
\label{sec:finetuning_cam}
In the main paper, we had demonstrated our method's target-edit-aware fine-tuning capabilities using the metadata-assisted UNet model. Here, we extend this experiment to the CAM~\cite{cam} framework. At capture time, CAM employes a learned sampler that generates a binary sampling mask $\mathbf{m}$ to select a set of RAW pixels. The mask $\mathbf{m}$ and the RAW samples $\mathbf{x_s}$ are saved as metadata, where the sample map $\mathbf{x_s}$ is computed by simply multiplying $\mathbf{m}$ with the RAW image $\mathbf{x}$ as $\mathbf{x_s} = \mathbf{m} \odot \mathbf{x}$, where $\odot$ denotes element-wise multiplication~\cite{cam}. During inference, the reconstruction network receives the sRGB image $\mathbf{y}$, the mask $\mathbf{m}$, and the RAW sample map $\mathbf{x_s}$ as input producing a reconstructed RAW image $\hat{\mathbf{x}} = f_\theta(\mathbf{y},\mathbf{m}, \mathbf{x_s})$. In the original CAM implementation, the fine-tuning loss is computed in RAW space over the sampled locations as:
\begin{equation}
\mathcal{L}_\mathrm{RAW-FT} =  \| \mathbf{m} \odot (\mathbf{x} - \hat{\mathbf{x}}) \|_2^2.
\end{equation}
We augment this with our edit-aware loss applied to the same sampled RAW values as:
\begin{equation}
\mathcal{L}_\mathrm{sRGB-FT} =  \| \mathbf{m} \odot (g_\phi(\mathbf{x}) - g_\phi(\hat{\mathbf{x}})) \|_2^2,
\end{equation}
where $g_\phi$, as described in the main paper, is our differentiable ISP with tunable parameters $\phi$.
Thus, the total fine-tuning objective becomes:
\begin{equation}
\mathcal{L}_\mathrm{total-FT} = \mathcal{L}_\mathrm{RAW-FT}
         + \lambda \mathcal{L}_\mathrm{sRGB-FT}.
\end{equation}

\begin{figure}
  \centering
  \includegraphics[width=\columnwidth]{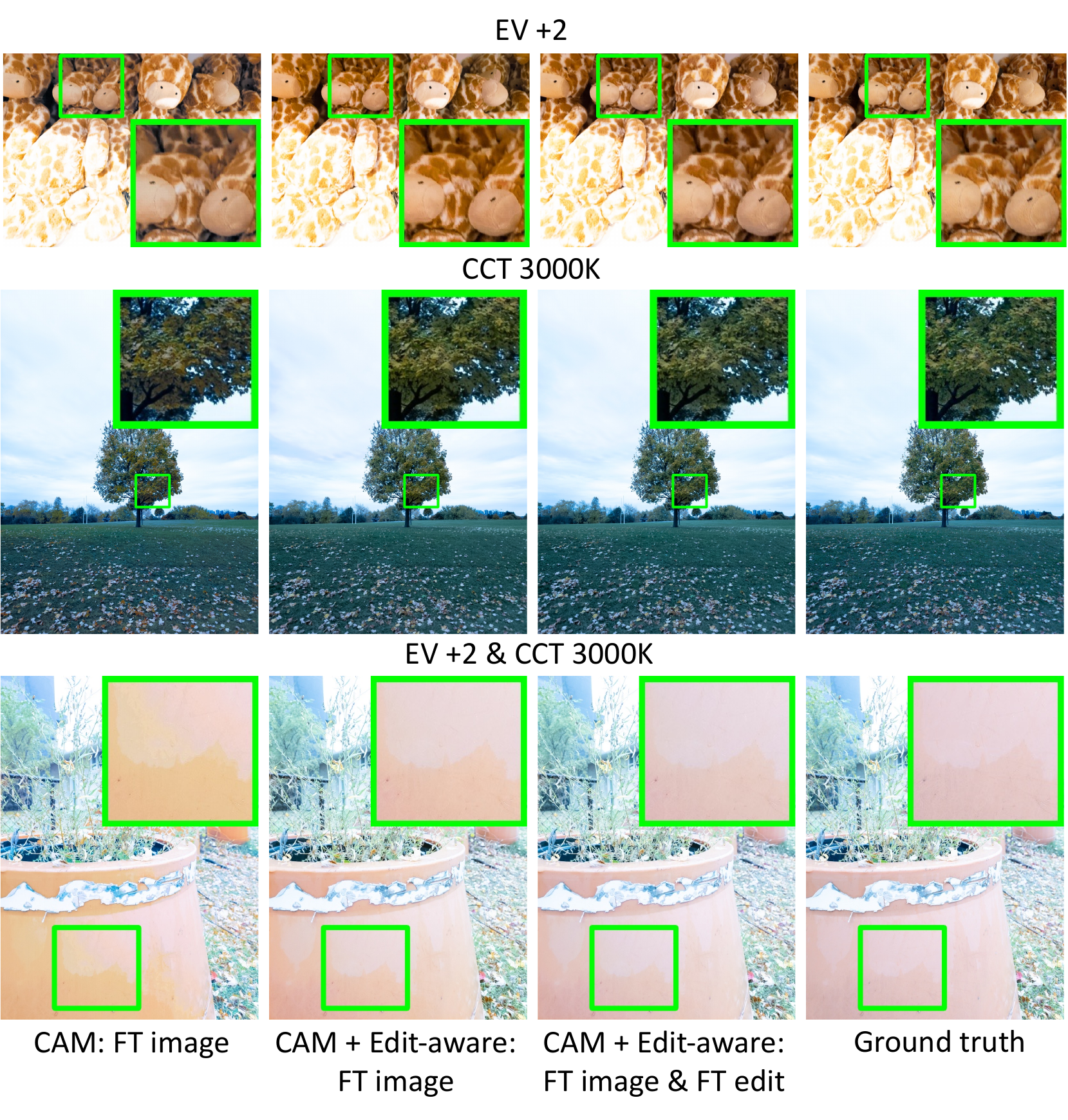}
  \caption{Examples of target-edit-aware fine-tuning applied to CAM~\cite{cam}. Standard CAM fine-tuning refines the reconstruction using only RAW-domain supervision, while our edit-aware objective allows the model to incorporate information about post-processing adjustments. When the fine-tuning edit is matched to the target adjustment, the sRGB rendering of the reconstructed RAW better aligns with the specified edit.
  }\label{fig:finetuning_supp}
%  \vspace{-5mm}
\end{figure}

Table~\ref{tab:finetuning_supp} reports results for the same three scenarios described in the main paper: (i) baseline CAM fine-tuned on the test image, (ii) our edit-aware model fine-tuned on the test image with randomly sampled $\phi$, and (iii) our model fine-tuned with $\phi$ fixed to the target edit. Even standard fine-tuning (scenarios i–ii) provides improvements over the baseline, while fixing $\phi$ to the target edit further enhances sRGB reconstruction fidelity. Qualitative examples in Fig.~\ref{fig:finetuning_supp} show that target-edit-aware fine-tuning produces outputs that are more consistent with the intended photofinishing adjustments.

\begin{table*}[]
\centering
\resizebox{0.775\textwidth}{!}{%
\begin{tabular}{l|lll|lll|lll}
\toprule
\multirow{2}{*}{Method}      & \multicolumn{3}{c|}{EV +2} & \multicolumn{3}{c|}{CCT 3000K} & \multicolumn{3}{c}{EV +2 \& CCT 3000K} \\
           & PSNR   & SSIM   & $\Delta$E   & PSNR    & SSIM   & $\Delta$E   & PSNR    & SSIM   & $\Delta$E   \\
\midrule
CAM: FT image       &           28.71   & 0.8750 & 4.39 & 29.11 & 0.8774 & 4.99 & 28.59 & 0.8725 & 4.47    \\

CAM + Edit-aware: FT image &            \rb{30.57} & \rb{0.8883} & \rb{3.73} & \rb{30.40} & \rb{0.8885} & \rb{4.34} & \rb{30.30} & \rb{0.8858} & \rb{3.83}    \\
CAM + Edit-aware: FT image \& FT edit     &   \yb{30.95} & \yb{0.8892} & \yb{3.49}  & \yb{30.67} & \yb{0.8896} & \yb{4.18}      & \yb{30.73} & \yb{0.8870} & \yb{3.61}  \\
\bottomrule
\end{tabular}
}
\caption{Quantitative results of fine-tuning (FT) the metadata-assisted CAM~\cite{cam} RAW reconstruction model. Results are averaged over the 400 test images from the RAW smartphone dataset of~\cite{s24}. Metrics include PSNR (dB), SSIM, and $\Delta$E for the edited sRGB renderings. EV denotes exposure value and CCT represents correlated color temperature. The \colorbox{yellow!35}{\textbf{best}} and \colorbox{red!15}{second-best} results are highlighted.
\label{tab:finetuning_supp}
}
\end{table*}

\section{Other edits}
\label{sec:other_edits}
In the main paper, we had showed that training with our lightweight differentiable ISP yields strong generalization to more sophisticated software ISPs, such as Adobe Photoshop. Although the tone and color operations implemented in Photoshop are substantially more complex than those represented in our pipeline, the model trained with our edit-aware loss produces consistent results under these transformations. In Table~\ref{tab:edits} of the main paper, the selected edits were aligned with the same broad class of operations modeled by our loss pipeline. To further examine generalization beyond these settings, we evaluate our method on two additional edits---dehazing and local tone mapping---that are not modeled in our loss formulation.

\begin{table}[]
\centering
\resizebox{\columnwidth}{!}{%
\begin{tabular}{l|lll|lll}
\toprule
\multirow{2}{*}{Method}         & \multicolumn{3}{c|}{Dehazing} & \multicolumn{3}{c}{Local tone mapping} \\
               & PSNR   & SSIM   & $\Delta$E   & PSNR  & SSIM   & $\Delta$E   \\
\toprule
CAM            &      26.08          &  0.8258   &  8.28  &  25.12  & 0.8514   & 8.31       \\
CAM + Edit-aware     &   \yb{27.71}             &  \yb{0.8565}  & \yb{6.22}   &  \yb{26.58}  & \yb{0.8692}    & \yb{6.15}        \\
\midrule
RAWDiff        &         23.49       &  0.8246   &  10.09  & 23.18   & 0.8604   & 12.14       \\
RAWDiff + Edit-aware &    \yb{24.59}            &  \yb{0.8445}  &  \yb{8.90}  & \yb{24.04}  & \yb{0.8638}   & \yb{10.60}       \\
\midrule
UNet           &      27.10          &  0.8376   &  \yb{6.62}   & 25.76   &  0.8531  &   6.93     \\
UNet + Edit-aware    &     \yb{27.12}           &  \yb{0.8547}   &  6.73   &  \yb{26.29}   & \yb{0.8675}   & \yb{6.25}       \\
\bottomrule
\end{tabular}
}
\caption{Quantitative evaluation of the proposed edit-aware loss integrated into different RAW reconstruction frameworks. Results are reported for CAM~\cite{cam}, RAW Diffusion~\cite{rawdiff}, and a UNet–based~\cite{unet} model, evaluated on 400 test images from the RAW smartphone dataset of~\cite{s24}. Metrics include PSNR (dB), SSIM, and $\Delta$E for the sRGB renderings. The \colorbox{yellow!35}{\textbf{best}} results are highlighted.
\label{tab:other_edits}
}
\end{table}

Dehazing is applied using Photoshop with the dehazing strength parameter in Adobe Camera RAW set to a value of 75. In the case of local tone mapping, we observed that Photoshop's output has relatively mild local tonal adjustments. To create a more challenging evaluation, we applied the contrast limited adaptive histogram equalization (CLAHE) algorithm~\cite{pizer1987adaptive}, which is a classical local tone mapping technique, on the output of Photoshop. In particular, we first rendered the RAW images to 16-bit uncompressed sRGB images using Photoshop under the Adobe Camera RAW default settings, and then applied the CLAHE local tone mapping algorithm on the results.

\begin{figure}[t!]
  \centering
  \includegraphics[width=\columnwidth]{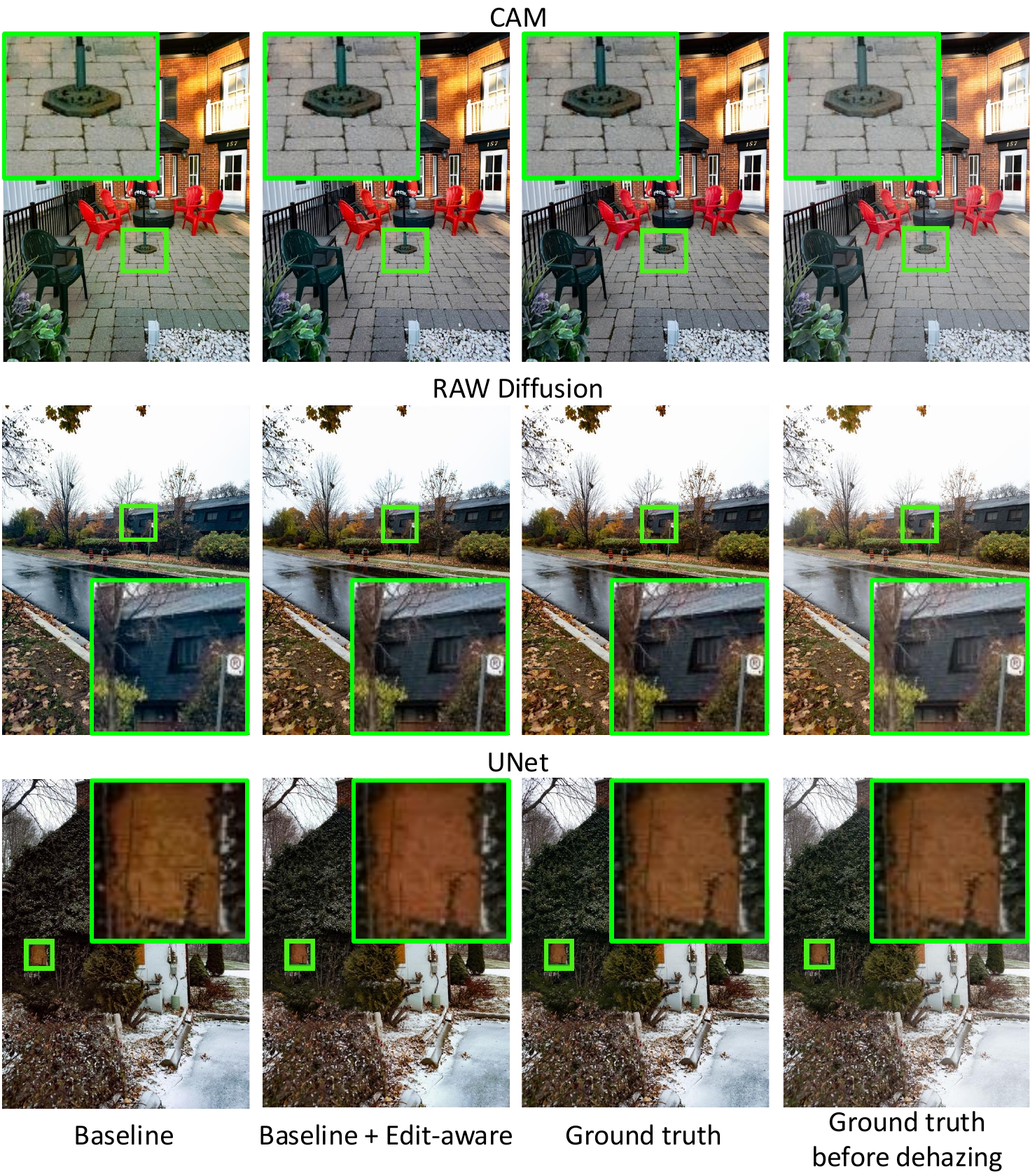}
  \caption{Dehazing examples with our edit-aware loss integrated into different RAW reconstruction frameworks. These results show that our approach generalizes effectively to a dehazing edit that was never included or approximated during training.
  }\label{fig:dehazing}
%  \vspace{-5mm}
\end{figure}

\begin{table}[]
\centering
\resizebox{0.375\columnwidth}{!}{%
\begin{tabular}{lc}
\toprule
\multirow{2}{*}{Model} &   sRGB \\         
   &   PSNR \\
\midrule
Fixed pipeline &         24.20    \\
Large sampling &         24.64    \\
\midrule
\textbf{Ours}        &        \textbf{25.15}  \\
\bottomrule
\end{tabular}
}
\caption{
Ablation study on sampling configurations. Results are reported using the UNet reconstruction model on a subset of 50 challenging images from the test split of the RAW smartphone dataset of~\cite{s24}. We report sRGB PSNR (dB) for Edit 5 (see Table~\ref{tab:edits} of the main paper for edit details). \label{tab:ablation_supp}
}
\end{table}

\begin{figure}
  \centering
  \includegraphics[width=\columnwidth]{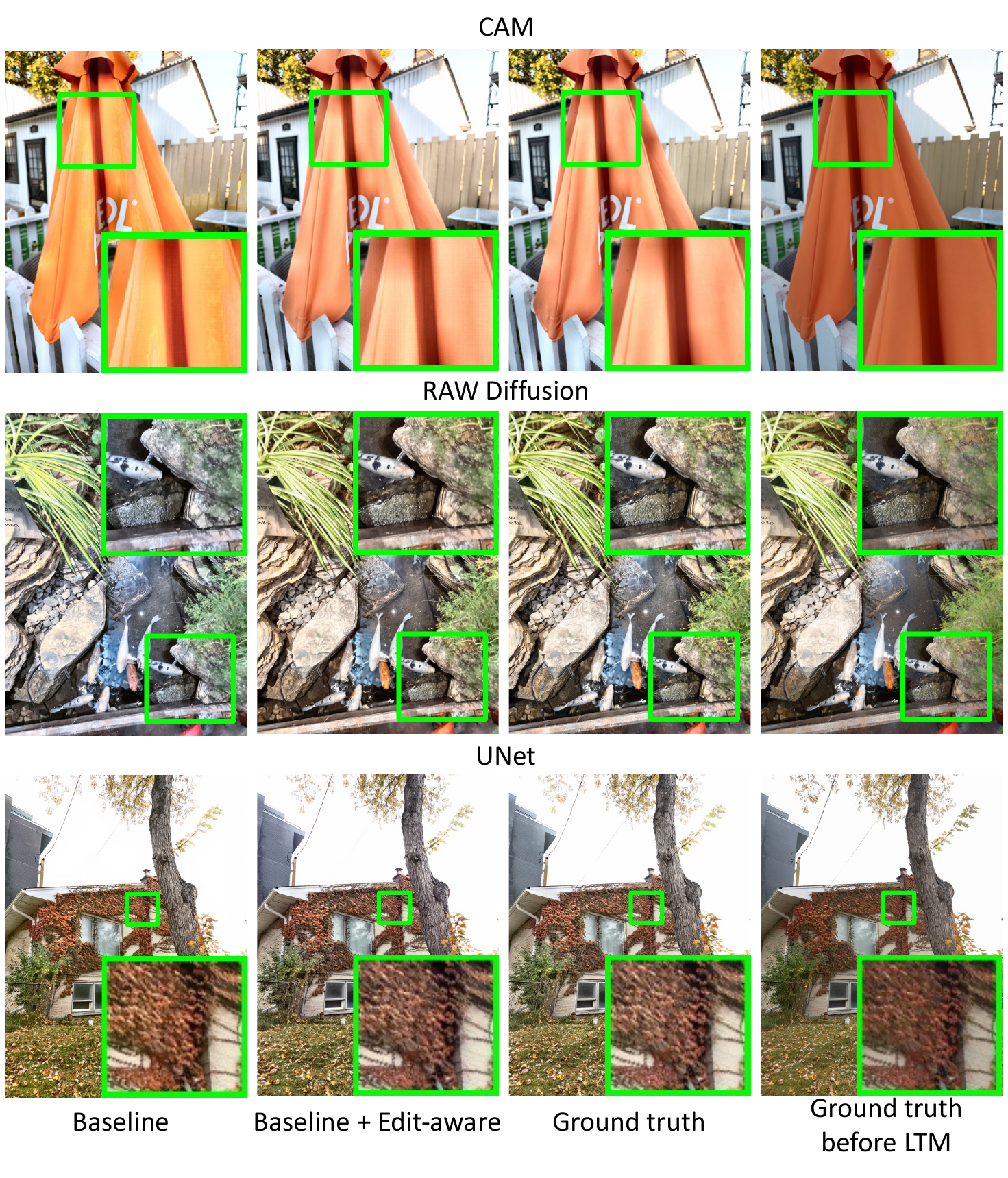}
  \caption{Local tone mapping (LTM) results with our edit-aware loss integrated into different RAW reconstruction methods. Despite not being explicitly modeled during training, spatially varying tone adjustments are handled effectively by our approach.
  }\label{fig:ltm}
%  \vspace{-5mm}
\end{figure}

Quantitative results for both edits are reported in Table~\ref{tab:other_edits}. Across all reconstruction frameworks, incorporating the edit-aware loss leads to improved performance, indicating that the proposed training strategy enhances robustness even for operations not explicitly modeled during training. Representative qualitative examples for dehazing and local tone mapping are shown in Figs.~\ref{fig:dehazing} and~\ref{fig:ltm}, respectively. 
The baseline reconstructions suffer from color distortions and inconsistent tones, while our method delivers outputs that align closely with the applied edits.

\section{Results on NUS~\cite{NUS} dataset}
\label{sec:nus}
We evaluated our method on three cameras from the NUS dataset~\cite{NUS} following the CAM~\cite{cam} training/testing protocol. Results for two edits are shown (Edit A: 1.5 EV, Cool, 3500K; Edit B: 2.5 EV, Blossom, As-shot WB). Identical loss hyperparameters as in Section~\ref{sec:results} and a fixed $\lambda=2$ (Eqn.~\ref{eqn:final_loss}) were used for all cameras.
As shown in Table~\ref{tab:nus_results}, our edit-aware method exhibits robust generalization across diverse cameras and edits. %These results will be included in the final version.

\section{Additional ablations}
\label{sec:additional_ablations}
Table~\ref{tab:ablation} of the main paper had examined the contributions of individual loss components and the effect of parameter sampling. In those experiments, using all loss terms but without sampling yielded weaker performance than our full configuration that includes randomized sampling of all edit parameters. It is equally important, however, that the sampled parameter ranges remain consistent with edits that users typically apply. For instance, our chosen configuration draws exposure adjustments from a normal distribution centered at zero and samples illuminants near the image’s ASN estimate, reflecting the fact that common edits tend to be moderate and anchored around default settings.

To further probe this sensitivity, we evaluate a setting in which the sampling distributions are broadened. Specifically, the exposure value is sampled from a uniform distribution $\varepsilon \sim \mathcal{U}(-3, 3)$; illuminant samples are drawn anywhere within the convex hull of the illuminant dictionary rather than being restricted to the neighborhood of the ASN ($l_1$ norm $\leq0.05$ was used as the threshold for all experiments); and tone mapping uses polynomials with higher degree $d=7$. These changes induce substantially stronger edits, and the resulting performance degradation is reflected in the second row of Table~\ref{tab:ablation_supp}. For ease of comparison, the fixed-pipeline baseline (sampling disabled) and our configuration are reproduced from Table~\ref{tab:ablation} of the main paper.

% Samsung NX2000, Olympus E-PL6, Sony SLT-A57
\begin{table}[!b]
% \vspace*{-0.325cm}
\centering
\resizebox{\columnwidth}{!}{%
\begin{tabular}{l|l|lll|lll}
\toprule
\multirow{2}{*}{Method}         & RAW       & \multicolumn{3}{c|}{Edit A} & \multicolumn{3}{c}{Edit B}  \\
               & PSNR & PSNR   & SSIM   & $\Delta$E   & PSNR  & SSIM   & $\Delta$E    \\
\toprule
Samsung            &   \yb{46.18}          &             32.86 & 0.9633 & 3.95 & 32.80 & 0.9708 & 3.05  \\
Samsung + EA     &    43.29       &             \yb{33.73} & \yb{0.9738} & \yb{2.75} & \yb{33.11} & \yb{0.9769} & \yb{2.53}     \\
\midrule
Olympus            &   \yb{50.01}          &             33.61 & 0.9533 & 4.13 & 33.42 & 0.9556 & 3.24  \\
Olympus + EA     &    48.76       &             \yb{35.88} & \yb{0.9699} & \yb{3.37} & \yb{34.66} & \yb{0.9656} & \yb{3.19}     \\
\midrule
Sony            &   \yb{51.02}          &             34.10 & 0.9573 & 4.05 & \yb{33.26} & 0.9582 & 3.94  \\
Sony + EA     &    49.28       &             \yb{34.55} & \yb{0.9657} & \yb{3.12} & \yb{33.26} & \yb{0.9626} & \yb{2.99}     \\
\bottomrule
\end{tabular}
}
\caption{
Results of CAM~\cite{cam} on three cameras (Samsung NX2000, Olympus E-PL6, Sony SLT-A57) from the NUS dataset~\cite{NUS}. Metrics include PSNR (dB), SSIM, and $\Delta$E for the edited sRGB renderings. The \colorbox{yellow!35}{\textbf{best}} results are highlighted. Our edit-aware method (EA) generalizes across cameras and edits. \label{tab:nus_results}
}
\vspace{-5mm}
\end{table}

Table~\ref{tab:lambda} shows an ablation on $\lambda$ on the challenging 50-image set in Section~\ref{subsec:ablations} of the main paper. It can be observed that performance is not highly sensitive to $\lambda$.
\begin{table}[!t]
\setlength{\tabcolsep}{2.5pt}
% \vspace*{-0.375cm}
\centering
\resizebox{\columnwidth}{!}{%
\begin{tabular}{l|lllll|lllll}
\toprule
Method  & \multicolumn{5}{c|}{CAM~\cite{cam}}       & \multicolumn{5}{c}{RAWDiff~\cite{rawdiff}}  \\
\cline{1-11}
     $\lambda$ value         & \multicolumn{1}{c|}{w/o EA} & \multicolumn{1}{c}{1}   & \multicolumn{1}{c}{2}   & \multicolumn{1}{c}{4} & \multicolumn{1}{c|}{8}   & \multicolumn{1}{c|}{w/o EA} & \multicolumn{1}{c}{1}   & \multicolumn{1}{c}{2}   & \multicolumn{1}{c}{4} & \multicolumn{1}{c}{8}    \\
\toprule
     sRGB PSNR          & \multicolumn{1}{c|}{22.4} & 23.7   & 24.6   & 23.9 & 23.8   & \multicolumn{1}{c|}{21.7}   & 22.9   & 23.2 & 23.5  & 23.0  \\
\bottomrule
\end{tabular}
}
\caption{
Ablation study on $\lambda$. Results are reported on a subset of 50 challenging images from the test split of the RAW smartphone dataset of~\cite{s24}. We report sRGB PSNR (dB) for Edit 5 (see Table~\ref{tab:edits} of the main paper for edit details). \label{tab:lambda}
}
% \vspace*{-0.45cm}
\end{table}

As mentioned in the main paper, certain blind methods, such as~\cite{nam2017modelling,ciexyznet,cycleisp,invisp,paramisp}, incorporate cyclic consistency constraints to encourage the reconstructed RAW to render back to the original sRGB input. This behaves similarly to using a fixed pipeline during training (row one of Table~\ref{tab:ablation_supp}), which we observed to be less effective when evaluating under diverse edits. To further scrutinize our edit-aware loss, we conduct an experiment using a forward–inverse network that incorporates a cyclic consistency objective. For this study, we adopt the CIE XYZ Net~\cite{ciexyznet} architecture. CIE XYZ Net consists of two sub-networks, a reverse-rendering module and a forward-rendering module, both jointly trained, where the forward-rendering sub-network is optimized using a supervised per-pixel sRGB fidelity loss to reproduce the input image. We first train a baseline model following this original setup. We then compare it against a variant in which we take only the reverse-rendering network and optimize it using the original per-pixel reconstruction loss and our edit-aware loss. As shown in Table~\ref{tab:cyclic_supp}, our edit-aware variant achieves substantially stronger performance under edits, highlighting that cyclic consistency alone does not encourage edit-compatible RAW recovery.

\begin{table}[]
\centering
\resizebox{0.7\columnwidth}{!}{%
\begin{tabular}{lccc}
\toprule
\multirow{2}{*}{Model} & \multicolumn{3}{c}{Edit 5} \\
& PSNR & SSIM & $\Delta$E \\         
   
\midrule
Cyclic loss &    20.54  &   0.7141 &  13.13   \\
\midrule
\textbf{Our edit-aware loss}        &  \textbf{27.13} & \textbf{0.8010} & \textbf{5.82} \\
\bottomrule
\end{tabular}
}
\caption{
Ablation study on using a cyclic consistency loss versus our edit-aware loss. Results are reported using the CIE XYZ Net~\cite{ciexyznet} model on a subset of 50 challenging images from the test split of the RAW smartphone dataset of~\cite{s24}. We report metrics for Edit 5 (see Table~\ref{tab:edits} of the main paper for edit details). \label{tab:cyclic_supp}
}
\end{table}

{
    \small
    \bibliographystyle{ieeenat_fullname}
    \bibliography{main}
}


\end{document}